%% file: acl2023.tex
\pdfoutput=1

\documentclass[11pt]{article}

\usepackage[]{ACL2023}

\usepackage{times}
\usepackage{latexsym}

\usepackage[T1]{fontenc}

\usepackage[utf8]{inputenc}

\usepackage{microtype}

\usepackage{inconsolata}

\def\ie{{\em i.e.,~}}
\def\eg{{\em e.g.,~}}

\usepackage{arydshln}
\usepackage{xspace}
\usepackage{subcaption}
\usepackage{mathtools,amssymb}
\usepackage{graphicx}
\usepackage{comment}
\usepackage{multirow}
\usepackage{amsmath, amsmath}
\usepackage{booktabs}
\usepackage{enumitem}
\usepackage{wrapfig}
\usepackage{cuted}

\usepackage{listings}
\usepackage{adjustbox}
\usepackage[normalem]{ulem}

\definecolor{dark-gray}{gray}{0.85}
\definecolor{light-gray}{gray}{0.95}
\definecolor{mygreen}{rgb}{0,0.4,0}
\definecolor{mygray}{rgb}{0.5,0.5,0.5}
\definecolor{mymauve}{rgb}{0.58,0,0.82}
\definecolor{myred}{rgb}{0.82, 0.1, 0.26}

\lstdefinestyle{CustomPy}{
    escapeinside={(*@}{@*)},
    belowcaptionskip=1\baselineskip,
    xleftmargin=1pt,
    xrightmargin=3pt,
    language=Python,
    numbersep=5pt,
    tabsize=4,
    showstringspaces=false,
    basicstyle=\scriptsize\ttfamily, 
    keywordstyle=\color{mygreen},
    commentstyle=\color{purple},
    stringstyle=\color{red},
    identifierstyle=\color{black},
    numberstyle=\tiny\color{mygray},
    emph={int,char,double,float,unsigned,void,bool,boolean},
    emphstyle={\color{myred}},
    emph=[2]{and, in,},
    emphstyle=[2]{\color{violet}},
    emph=[3]{sortedCount, sorted_count},
    emphstyle=[3]{\color{blue}},
    numbers=left,
    stepnumber=1,
    breaklines=true,
    backgroundcolor=\color{white},
}

\lstdefinestyle{CustomJava}{
    belowcaptionskip=1\baselineskip,
    xleftmargin=1pt,
    xrightmargin=3pt,
    language=Java,
    numbersep=5pt,
    tabsize=2,
    showstringspaces=false,
    basicstyle=\scriptsize\ttfamily, 
    keywordstyle=\color{mygreen},
    commentstyle=\color{purple},
    stringstyle=\color{red},
    identifierstyle=\color{black},
    numberstyle=\tiny\color{mygray},
    stringstyle=\color{mymauve},
    emph={int,char,double,float,unsigned,void,bool,boolean},
    emphstyle={\color{myred}},
    emph=[2]{and, in,},
    emphstyle=[2]{\color{violet}},
    emph=[3]{sortedCount, sorted_count},
    emphstyle=[3]{\color{blue}},
    numbers=left,
    stepnumber=1,
    breaklines=true,
    backgroundcolor=\color{white},
    literate={\ \ }{{\ }}1,
}

\newcommand{\ourmethod}{\textsc{ContraCLM}\xspace}
\newcommand{\ourmethodtok}{\textsc{ContraCLM-Tok}\xspace}
\newcommand{\ourmethodseq}{\textsc{ContraCLM-Seq}\xspace}

\usepackage{titlesec}
\titlespacing{\paragraph}{%
  0pt}{
  0.2\baselineskip}{
  1em}%

\title{\ourmethod: Contrastive Learning For Causal Language Model}

\author{Nihal Jain$^\ast$,   Dejiao Zhang\thanks{~ Equal Contribution. Accepted to ACL 2023. Correspondence to Dejiao Zhang $<$dejiaoz@amazon.com$>$.}, Wasi Uddin Ahmad$^\ast$, Zijian Wang, Feng Nan, \\ \textbf{Xiaopeng Li},
\textbf{Ming Tan}, \textbf{Ramesh Nallapati}, \textbf{Baishakhi Ray}, \\ \textbf{Parminder Bhatia}, \textbf{Xiaofei Ma}, \textbf{Bing Xiang}  \\
AWS AI Labs,  USA \\
}

\begin{document}
\maketitle

\input{sections/0.abstract}
\input{sections/1.introduction}
\input{sections/2.related_work}
\input{sections/3.method}

\input{sections/4.0.experiments}

\input{sections/4.2.eval_nl}

\input{sections/4.3.eval_pl}

\input{sections/4.4.ablation}
\input{sections/5.conclusion.tex}


\subsection*{Author Contributions}
Dejiao and Wasi proposed the initial framework for \ourmethod and completed the paper writing. Nihal and Dejiao set up the pretraining code. Nihal processed the pretraining data for the programming language. Dejiao designed and completed all natural language related training and evaluations. Nihal and Wasi complete the associated counter parts for the programming data.  Zijian is in charge of the pretraining data collection and multinode distributed training of ContraGen models on the programming data.  Feng and Xiaopeng helped on some preliminary explorations on natural language evaluation. All the other co-authors provided thought-provoking discussions and suggestions during the weekly meeting for this project and helped shape and proofread the paper draft.

\subsubsection*{Acknowledgments}
We thank all the helpful discussions and comments from colleagues at AWS AI Labs. 

\bibliography{custom}
\bibliographystyle{acl_natbib}

\clearpage
\appendix
\input{appendix}

\end{document}

%% file: sections/0.abstract.tex
\begin{abstract}
Despite exciting progress in causal language models, 
the expressiveness of the representations is largely limited due to poor discrimination ability.
To remedy this issue, we present \ourmethod, a novel contrastive learning framework at both token-level and sequence-level. We assess \ourmethod on a variety of downstream tasks. We show that \ourmethod enhances discrimination of the representations and bridges the gap with the encoder-only models, which makes causal language models better suited for tasks beyond language generation. Specifically, we attain $44\%$ relative improvement on the Semantic Textual Similarity tasks and $34\%$ on Code-to-Code Search tasks. Furthermore, by improving the expressiveness of the representations, \ourmethod also boosts the source code generation capability with $9\%$ relative improvement on execution accuracy on the HumanEval benchmark.
\end{abstract}

%% file: sections/1.introduction.tex
\section{Introduction}
\label{sec:intro}
Causal Language Models (CLM) have seen remarkable success in language generation, both in natural language \citep{gpt1, gpt2, gpt3} and programming language \citep{codex, codegen}.
However, one limitation at its core is the poor discrimination ability of the representations, which often causes a large performance gap with the encoder-only or encoder-decoder models on discriminative tasks (see Appendix \ref{appendix:bridge_the_gap}) and hence limits the wide usage of CLM beyond language generation.

\begin{figure*}
     \centering
     \begin{subfigure}[b]{0.48\textwidth}
         \centering
         \includegraphics[width=0.98\textwidth]{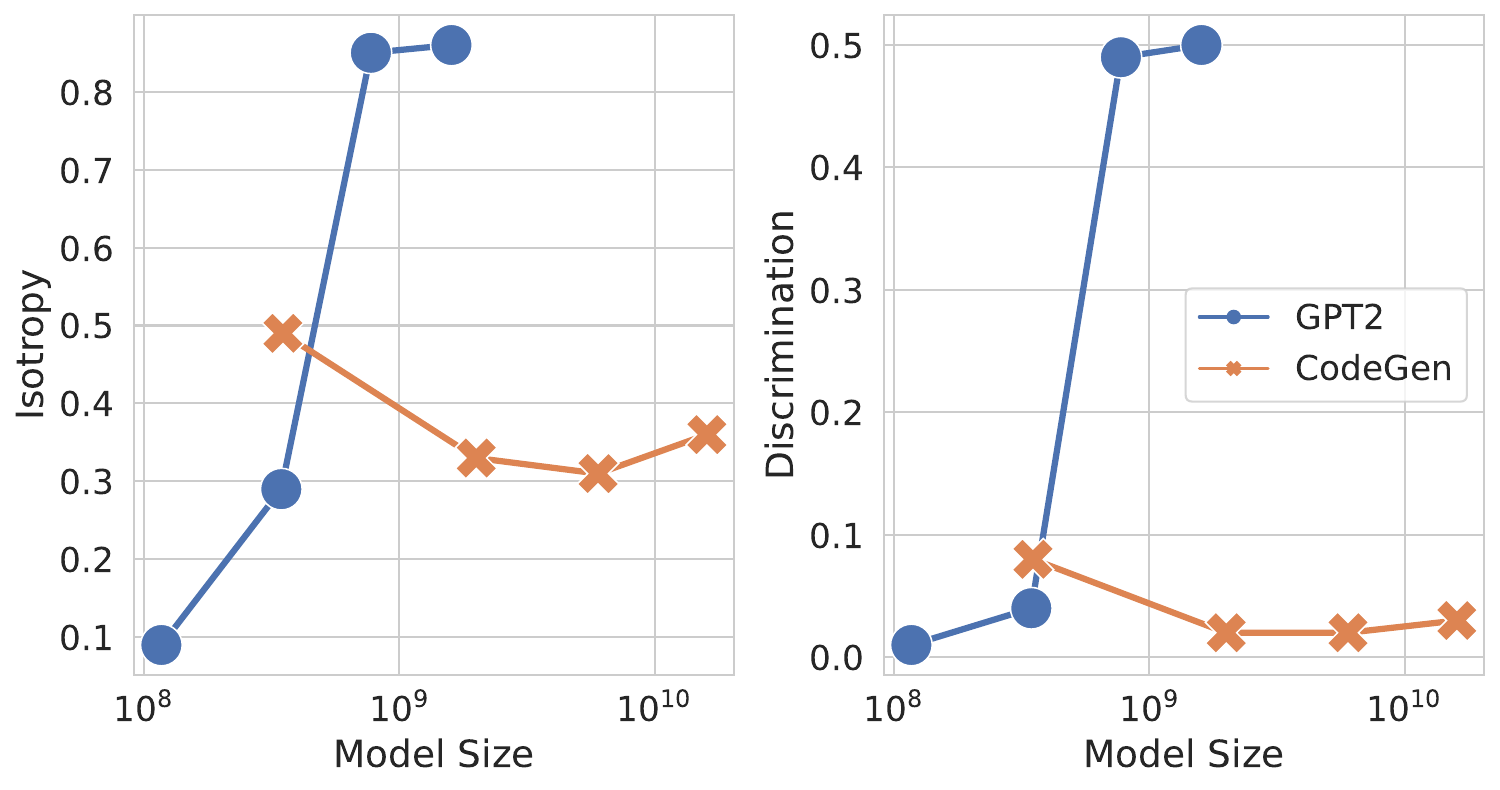}
         \caption{The isotropy and discrimination abilities of the representations yielded by a model do not always improve with the increased model size.}
         \label{fig:variant_scaling_law}
     \end{subfigure}
     \hspace{0.2cm}
     \begin{subfigure}[b]{0.48\textwidth}
         \centering
         \vspace{2mm}
         \includegraphics[width=0.98\textwidth]{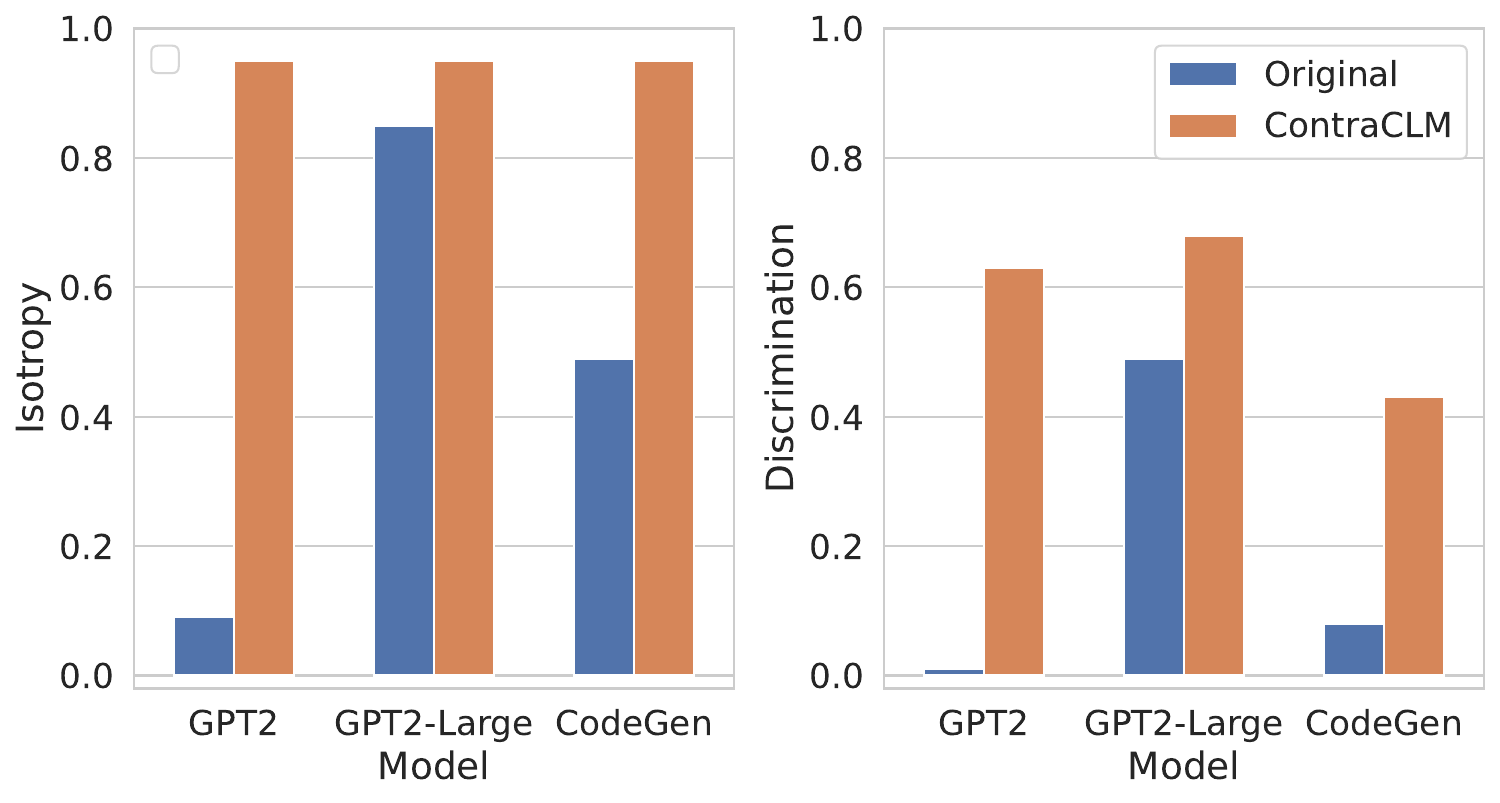}
         \caption{\ourmethod can effectively enhance both isotropy and discrimination, regardless of whether the original models suffer from the degenerated representations or not. }
         \label{fig:effective_contraclm}
     \end{subfigure}
    \vspace{-1mm}
    \caption{Evaluating the representation quality regarding both $\underline{\text{isotropy}} := 1 - \textit{inter-similarity}$ and $\underline{\text{discrimination}} := 1 - \textit{intra-similarity / inter-similarity}$, where \textit{intra-similarity} refers to the average cosine similarity between tokens from the same sequence and inter-similarity is defined regarding tokens from two randomly sampled sequences. We use WIT \citep{witdata} and code search dataset \cite{guo-etal-2022-unixcoder} to evaluate GPT2 and CodeGen models.}
    \vspace{-2mm}
    \label{fig:motivation}
\end{figure*}

Prior studies posit the \textit{anisotropy} issue, \ie representations being squeezed into a tiny cone in the vector space \citep{contextual_word_embeddings}, can be the main cause for poor discrimination ability of language models across different architectures and objectives. Many efforts have focused on resolving the anisotropy issue on encoder-only or encoder-decoder models, either through post-processing \citep{bertwhitening, bertflow} or integrating different regularization terms into the training objective \citep{cosreg, spectrumreg}. 
A recent work \citep{contrastive-search} shows that the decoder-only CLM does not suffer from the anisotropic problem as long as the model is beyond a certain size. However, we find such conclusions can vary across domains. As shown in Figure \ref{fig:variant_scaling_law}, CLMs pretrained on text, \ie GPT-2 \citep{gpt2}, do yield representations with good isotropy and discrimination as long as the model is not smaller than 774M parameters (GPT2-Large), whilst CodeGen \citep{codegen}, pretrained on the programming data, consistently suffers from anisotropy and poor discrimination across different model sizes. Therefore, an effective training strategy is still essential for CLMs to improve representation quality with better isotropy and discrimination (Figure \ref{fig:effective_contraclm}).
We conjecture that not only for models suffering from inferior representations, \eg CodeGen, and GPT2 (124M) but also for those with a good starting point (suffer less \eg GPT2-Large (774M)). 


We argue an ideal CLM should yield isotropic representations to leverage the representation space better, as well as discriminative representations such that tokens or sequences from the same context are mapped to comparatively closer locations in the vector space compared to those from randomly sampled contexts. To this end, we developed \ourmethod, a novel contrastive learning framework at both token-level and sequence-level. 
\ourmethod is able to promote more uniformly distributed and hence isotropic representations by separating the instances at different semantic levels, \eg tokens or sequences, apart from each other. \ourmethod improves the discrimination of representations due to the implicit grouping effect on semantically similar instances, yielded by pulling together the semantic-preserved variations, a.k.a. positive pair, of the same instance \citep{uniformvsalignment,wang2021understanding,zhang2021supporting}.


A natural question arises as to how would the improved representations affect the generation ability of CLMs.  Towards addressing this, we assess \ourmethod on language generation tasks in different domains, where we achieve better MAUVE \citep{pillutla2021mauve} on text generation and $9\%$ relative improvement on $\text{pass@}1$ accuracy on HumanEval \citep{codex}. The improvement in code completion is indeed significant as it reflects that more model-generated programs pass a suite of test cases. 
On the discriminative tasks, \ourmethod attains $44\%$ relative improvement on Semantic Textual Similarity tasks and $34\%$ on Code-to-Code Search tasks, which largely bridges the gap with the encoder-only or encoder-decoder models (see Section \ref{sec:ablation} and Appendix \ref{appendix:bridge_the_gap}). Such improvements allow more potential to leverage decoder-only models, especially considering their rapidly increasing sizes, to further boost their performance on a wide range of discriminative tasks where encoder-based models are currently the workhorse.

%% file: sections/2.related_work.tex
\section{Related Work}
\label{sec:relate_work}
\paragraph{Anisotropic Representation of Language Models} Despite the remarkable success achieved by language models \citep{devlin2018bert, gpt2, yang2019xlnet, textt5, textbart}, they suffer from the \textit{anisotropy} issue where the representations are distributed into a tiny cone in the vector space \citep{cosreg, contextual_word_embeddings, bertflow, spectrumreg}. 
In particular, \citet{contextual_word_embeddings} shows that the degeneration is severer on CLM, where the average cosine similarity between two words sampled from randomly selected sequences is almost at one when evaluating the outputs from the last hidden layer of GPT-2 \citep{gpt2}. 
However, \citet{contrastive-search} show that CLMs \citep{gpt2} are indeed coherent as long as the model is larger than a certain size. We find such conclusions can vary across domains, \eg when pretraining on code, CodeGen \citep{codegen} consistently suffers from the anisotropy issue over a wide range of model sizes. 
On the bright side, Figure \ref{fig:effective_contraclm} shows that \ourmethod can effectively improve the representation quality when we continue to train the existing CLMs with our proposed objectives, regardless of whether the CLMs suffer from inferior representations initially.



\paragraph{Contrastive Learning}
Contrastive learning \citep{chen2020simple,he2020momentum} has seen remarkable successes in Natural Language Processing (NLP). A large amount of research has focused on sentence representation learning for encoder-only models, with the main differences lie in how the augmentations are generated \citep{fang2020cert,giorgi2020declutr,wu2020clear,meng2021coco,yan2021consert,kim2021self,gao2021simcse,vascl}. Recently there is an emerging interest in developing effective contrastive learning approaches for text generation models. However, most existing work mainly focuses on the encoder-decoder structure \citet{unilm, textt5, textbart} by contrasting suboptimal model generations obtained via diverse sampling \citep{an2022cont} or adding perturbations on the embedding space \citep{lee2021contrastive}, against the ground truth. On the other hand, it is not intuitive to develop an effective contrastive learning strategy for the decoder-only models. A recent work \citep{simctg} proposes SimCTG, a token-level contrastive learning approach that aims to separate each token apart from others within the same sequence by a prefixed distance. As shown in Section \ref{sec:experiments}, our temperature-based token-level contrastive learning approach, \ourmethodtok, consistently outperforms SimCTG across different tasks. We conjecture that the fixed margin-based objective allows less flexibility for the token-level representation separation, especially considering how the semantic relevance among tokens can vary across contexts (sequences). 


\paragraph{Code Generation and Beyond} Language modeling for source code is a fast growing area of research. Various model architectures have been explored recently, including  encoder-only \citep{codebert, graphcodebert}, encoder-decoder \citep{plbart, codet5, alphacode}, and decoder-only models \citep{codex, codegen, palm}. 
Among them, the decoder-only models were found effective on the code generation front. However, as shown in Section \ref{sec:code_completion} and Appendix \ref{appendix:bridge_the_gap}, they suffer from unsatisfactory performance on various discriminative tasks \citep{codexglue, huang-etal-2021-cosqa, guo-etal-2022-unixcoder}.
This motivates us to improve the decoder-only models on the discriminative tasks so as to extend their main usage beyond language generation. Furthermore, code is fundamentally different from natural language in that it is more structured, which helps validate the generalizability of our approach beyond plain text. 



%% file: sections/3.method.tex
\section{Model}
\label{sec:model}

\subsection{Causal Language Modeling}
Let $\mathbf{x} = [\mathbf{x}_1, \mathbf{x}_2, \cdots, \mathbf{x}_{|\mathbf{x}|}]$ denote a sequence with variable length $|\mathbf{x}|$, \eg a piece of text or a code snippet.
Causal Language Modeling (CLM)
is usually formulated as sequence distribution estimation over a set of sequences, $\mathbf{x}^1, \mathbf{x}^2, \dots, \mathbf{x}^N$. 
For tractable estimation, common practice is to factorize the joint distribution of each sequence into the product of conditional token prediction probabilities. The model is then trained via maximum likelihood estimation as 
follows,
\begin{align}
    \mathcal{L}_{\text{CLM}} = - \frac{1}{N}\sum_{j=1}^{N}\sum_{i=1}^{|\mathbf{x}^j|} \log p(\mathbf{x}_i^j | \mathbf{x}_{< i}^j) \;. 
    \label{eq:clm_loss} \nonumber
\end{align}
Here $\mathbf{x}_{< i}^j = [\mathbf{x}_1^j, \ldots, \mathbf{x}_{i-1}^j]$ denotes the subsequence before $\mathbf{x}_{i}^j$ and $|\mathbf{x}^j|$ is the sequence length. 


\subsection{Contrastive Learning for CLM}
Let $\mathbf{h}^{(i)}, \mathbf{h}^{(i^+)}$  denote two semantic-preserved representation variations of the same instance, a.k.a. the positive pair for contrastive learning. Then denote $\mathcal{I} = \{1, 2, \dots, N\} \cup \{1^+, 2^+, \dots, N^+\}$ as the set of representation indices associate with $N$ instances. Further, let $\tau$ denote the temperature hyper-parameter, and $\diamond$ denote the cosine similarity.  We then minimize the following, 
\begin{equation}
\label{eq:unified_cl_objective}
\small
\begin{split}
    \mathcal{L} = \sum_{j = 1}^{N}  - \left(\log \frac{\exp(\mathbf{h}^{(j)} \diamond \mathbf{h}^{(j^+)}/\tau)}{\sum_{k \in \mathcal{I} \setminus j } \exp(\mathbf{h}^{(j)} \diamond \mathbf{h}^{(k)}/\tau)} \right. \\
    \left. + \log \frac{\exp(\mathbf{h}^{(j^+)} \diamond \mathbf{h}^{(j)}/\tau)}{\sum_{k \in \mathcal{I} \setminus j^+ } \exp(\mathbf{h}^{(j^+)} \diamond \mathbf{h}^{(k)}/\tau)} \right)
    \;.
\end{split} \nonumber
\end{equation}
Note that in our setting, an instance can refer to either a token or a sequence. When $\mathbf{h}^{(j)}, \mathbf{h}^{(j+)}$ denotes the representations of the $j$-th token within a sequence, $N$ is the sequence length that can vary across sequences. For the sequence-level contrastive loss, $\mathbf{h}^{(j)}, \mathbf{h}^{(j+)}$ refer to the hidden representations of the $j$-th sequence within a batch, and $N$ denotes the batch size.

Therefore, when applied at both token-level and sequence-level\footnote{Please refer to Appendix~\ref{appendix:model} for the formulations.}, the contrastive learning objective defined above tries to separate tokens at each distinct location apart from every other token within the same sequence, and sequences within the same randomly sampled batch apart from each other.  Intuitively, such separation can improve the uniformity (isotropy) of the representations. Further, better discrimination of the representations is achieved due to the implicit grouping effect of contrastive learning on semantically similar instances, yielded by the explicit force to bring the semantic-preserving variations of the same instance together in the representation space. Such grouping effect of contrastive learning has been studied in recent work \citep{wang2021understanding, zhang2021supporting, uniformvsalignment} as well.  


\subsection{\ourmethod}
\label{sec:contraclm_model}
In addition to the causal language modeling loss, \ourmethod optimizes the contrastive learning objective defined in Equation \eqref{eq:unified_cl_objective} at both the token-level ($\mathcal{L}_{\text{Tok}}$) and sequence-level ($\mathcal{L}_{\text{Seq}}$) as follows
\begin{equation}
\label{eq:contraclm_objective}
\begin{split}
    \mathcal{L}_{\ourmethod} = \mathcal{L}_{\text{CLM}} + \mathcal{L}_{\text{Tok}} + \mathcal{L}_{\text{Seq}} \;.
\end{split}\nonumber 
\end{equation}
Furthermore, to understand how the token-level and sequence-level contrastive learning contribute to the overall performance, we assess the performance of  $\mathcal{L}_\ourmethodtok = \mathcal{L}_{\text{CLM}} + \mathcal{L}_{\text{Tok}}$ and $\mathcal{L}_\ourmethodseq = \mathcal{L}_{\text{CLM}} + \mathcal{L}_{\text{Seq}}$ in Section \ref{sec:experiments}. 
Unless otherwise specified, we weigh each loss equally and set the temperature $\tau=0.05$. 
Although better performance can be achieved by hyperparameter optimization, we mainly investigate how \ourmethod improves the representation quality and the zero-shot transfer learning performance. We hence leave hyperparameter optimization in a supervised setting as future work.

\paragraph{Positive pair of representations} 
For GPT-2 \citep{gpt2}, we consider the simple yet effective dropout-based augmentation \citep{gao2021simcse}, where the positive representation pair is obtained by forwarding the sequence twice. On the other hand, for CodeGen \citep{codegen}, we simply duplicate the representation of each instance as positive pair for an apples-to-apples comparison since dropout is disabled during its initial pretraining stage.
Unlike the existing findings that the dropout-based augmentation can boost the contrastive learning performance when (continually) training a language model, we find that the trends can vary when evaluating the discrimination tasks and the generation tasks. Detailed ablation study can be found in Section \ref{sec:ablation} and Appendix \ref{appendix:dropout_ablation}.

%% file: sections/4.0.experiments.tex
\section{Experiments}
\label{sec:experiments}
To demonstrate the effectiveness of our proposed framework in different application domains, we evaluate our models and baselines on natural language and programming language tasks. 
We design our experiments to address -- (1) Does contrastive learning improve the \emph{discrimination ability} of representations? (2) Do the representations learned by contrastive learning lead to better performance on \emph{language generation} tasks? (3) Is the joint contrastive learning at both token- and sequence-level necessary, and how do they benefit from each other? (4) How does the impact of contrastive learning vary across language domains?


\subsection{Data and Models}
\label{sec:data_and_model}
\paragraph{Data \& Models} For text, we continue training GPT-2 (124M) \citep{gpt2} on  WikiText-103, a collection of over 100 million tokens extracted from the set of verified Good and Featured articles on Wikipedia \citep{merity2016pointer}. For code, we continue training CodeGen 350M monolingual \citep{codegen} on collected permissively licensed Python code from GitHub.  Please refer to Appendix \ref{appendix:training} for the training details. We consider the following objectives for the continual training of both GPT-2 and CodeGen:

\begin{itemize}[noitemsep,nolistsep,leftmargin=*]
    \item \textbf{CLM}. The standard left-to-right autoregression objective for training causal language models, which is also the objective used for pretraining both GPT-2 and CodeGen. 
    \item \textbf{SimCTG} \citep{simctg}. A predefined margin\footnote{For all experiments in this section, we set the margin $\rho=0.5$ as recommended in \citet{simctg}.} based token-level contrastive learning framework that aims to separate tokens at each distinct location within a sequence apart from each other.
    \item \textbf{\ourmethodtok} \& \textbf{\ourmethodseq}. 
    As defined in Section \ref{sec:contraclm_model}, these two are obtained by combining the CLM objective with our proposed token-level or sequence-level contrastive loss, respectively. The investigation allows us to better understand how our token-level and seqeunce-level contrastive losses contribute to the overall performance of \ourmethod.
\end{itemize}

%% file: sections/4.2.eval_nl.tex
\subsection{Evaluation on Natural Language}
\label{sec:nl_experiment}
We first evaluate our model on discrimination and generation tasks in natural language. 

\begin{table*}[htbp]
    \begin{center}
    \resizebox{0.9\textwidth}{!}{
    \begin{tabular}{rcccccccc}
    \toprule
    \text{Model}& \text{STS12} & \text{STS13} & \text{STS14} & \text{STS15} & \text{STS16} & \text{SICK-R} & \text{STS-B} & \text{Avg.} \\ 
    \cmidrule(l){1-9} 
    GPT2&     25.84	& 28.90 &	26.20 &	34.74 &	35.70 &	42.72 &	26.27 &	31.48 \\
    CLM &     27.14&	20.34&	18.73&	37.56&	27.40&	35.70&	27.97&	27.83 \\
    SimCTG &  30.32&	37.10&	31.99&	39.68&	42.73&	46.26&	25.27&	36.19 \\
  \cmidrule(l){1-9}
  
    \ourmethodtok & 37.28	&37.63	&31.33	&54.78	&50.16	&48.10	&34.95	&42.03 \\ 
    \ourmethodseq & 29.66   &39.89   &34.50   &43.20  &41.99           &44.52        &25.51    &37.04\\ 
    \ourmethod & \textbf{37.54}	  &\textbf{45.23}	&\textbf{36.41}	 &\textbf{56.74}   &\textbf{50.30}   &\textbf{51.52}    &\textbf{39.49}   &\textbf{45.32} \\
    \bottomrule
    \end{tabular}
    }
    \caption{Spearman rank correlation between the cosine similarity of sentence representation pairs and the ground truth similarity scores.
    }
    \vspace{-2mm}
    \label{tab:STS_compare}
  \end{center}
\end{table*}

\begin{figure*}[ht]
\begin{minipage}{0.5\textwidth}
    \centering
    \includegraphics[width=1.0\textwidth]{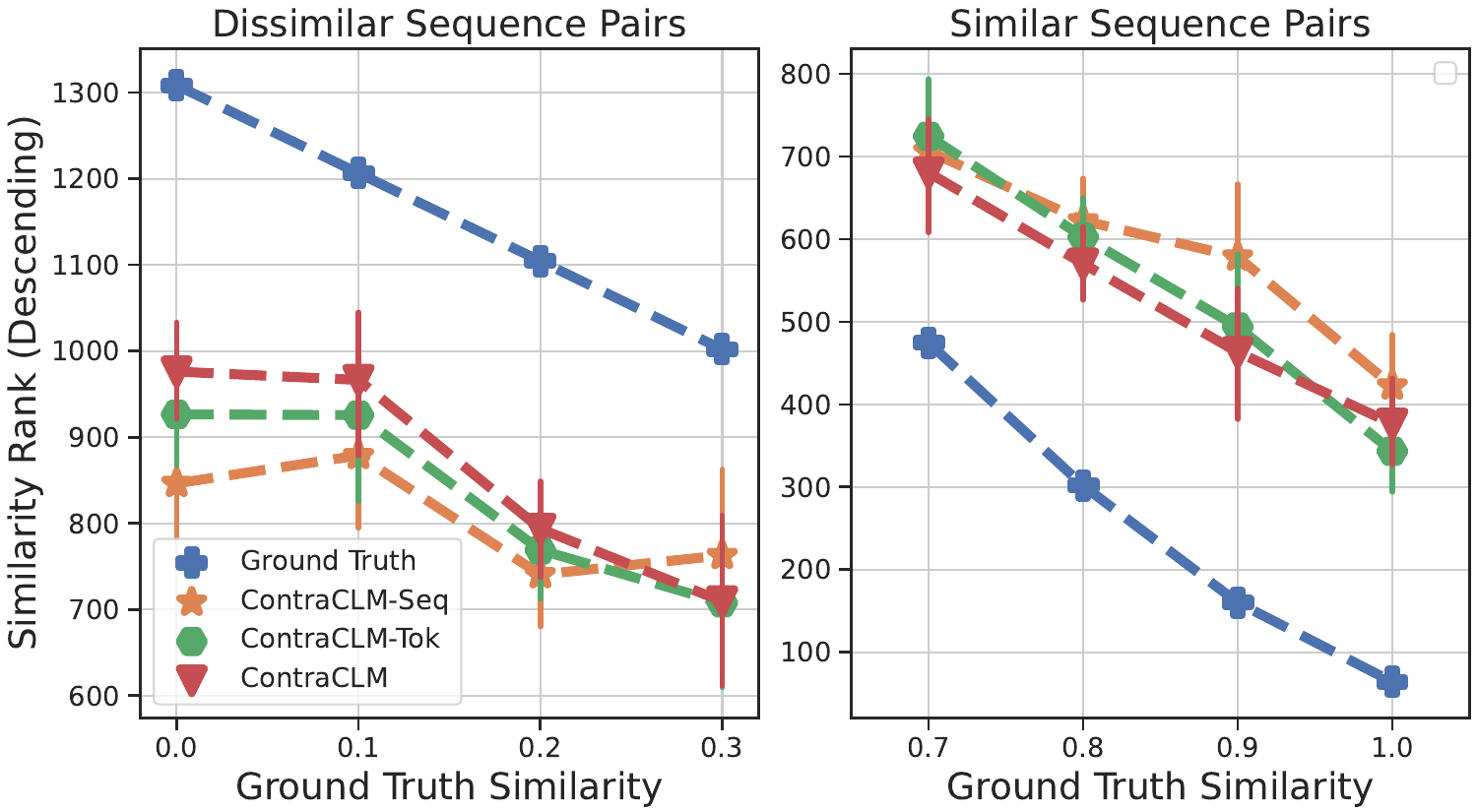}
\end{minipage}
\hspace{0.2cm}
\begin{minipage}{0.46\textwidth}
    \centering
    \resizebox{1.0\textwidth}{!}{%
    \setlength{\tabcolsep}{3pt}
    \def\arraystretch{1.0}%
    \begin{tabular}{l l c }
    \toprule
    \textbf{Similar Sequence Pair}&  Model & Rank$\downarrow$    \\
    \hline
    \multirow{2}{*}{\shortstack[l]{\textbf{S1:} a woman is stabbing \\ a potato with a fork}}
    &Ground Truth        &40    \\
    &\ourmethodseq      &501       \\
    \multirow{2}{*}{\shortstack[l]{\textbf{S2:} a woman is punctu- \\ ring a potato with a fork}} 
    &\ourmethodtok     &272       \\  
    &\ourmethod         &\textbf{251}       \\ 
    \midrule
    \textbf{Dissimilar Sequence Pair}&  Model & Rank$\uparrow$ \\
    \midrule
    \multirow{2}{*}{\shortstack[l]{\textbf{S1:} a man is opening a \\ 
    box and taking out paper}}
    &Ground Truth        &1310    \\
    &\ourmethodseq      &400     \\ 
    \multirow{2}{*}{\shortstack[l]{\textbf{S2:} a woman is peeling \\ a  
     potato}}
    &\ourmethodtok     &1054     \\
    &\ourmethod         &\textbf{1181}   \\ 
    \bottomrule                         
    \end{tabular}%
}
\end{minipage}
\caption{
\ourmethodtok is essential for making the sequence-level representations robust to spurious patterns of words or phrases (results reported on STS-B). We scale the ground truth similarity scores from [0, 5] to [0,1].
}
\vspace{-2mm}
\label{fig:contragen_tok_effect}
\end{figure*}

\subsubsection{Semantic Textual Similarity}
We assess \ourmethod on semantic textual similarity (STS), the most commonly used benchmark for evaluating the semantic discrimination capability of representations. STS consists of seven tasks, namely STS 2012-2016 \citep{agirre2012semeval,agirre2013sem,agirre2014semeval,agirre2015semeval,agirre2016semeval}, the STS Benchmark \citep{cer2017semeval}, and the SICK-Relatedness  \citep{marelli2014sick}. Human annotators provide a fine-grained similarity score from 0 to 5 for each sequence pair. Following \citet{sentence-bert}, for the sequence pairs in each dataset, we report the overall Spearman’s correlation between the cosine similarities of representations and the human-provided similarity scores in Table \ref{tab:STS_compare}.

\paragraph{Effectively Enhancing Discrimination} 
Table \ref{tab:STS_compare} shows that both GPT-2 and the one continually trained with CLM perform poorly on STS, which is a consequence of poor discrimination: the cosine similarities between semantically similar or dissimilar pairs are both almost at one (Figure \ref{fig:appendix_sts_rank_and_score_analysis} in Appendix \ref{appendix:sts_analysis}).
Also note that continuing to train GPT-2 with CLM on WikiText-103 worsens performance, which can occur since  WikiText-103 and the domain of the STS datasets is different.\footnote{STS datasets include text from image captions, news headlines and user forums. As a result, adapting GPT-2 to WikiText-103 reduces its transfer ability.}
In contrast, both \ourmethod and SimCTG largely outperform GPT-2, yet still, \ourmethod attains $25\%$ relative improvement over SimCTG. Moreover,  \ourmethodtok outperforms SimCTG on almost all STS benchmarks and the trend remains the same even without the dropout-based augmentation (Appendix \ref{appendix:simctg_vs_contragen}).
Therefore, we posit our temperature-based contrastive learning objective allows more flexibility in semantic-dependent separations among tokens while requiring a prefixed separation margin between tokens (as SimCTG does) is not ideal.

\paragraph{\ourmethodtok vs. \ourmethodseq} 
Table \ref{tab:STS_compare} also indicates that \ourmethodtok and \ourmethodseq complement each other, as \ourmethod consistently performs better than both of them on STS. Note that \ourmethodseq performs worse than \ourmethodtok. It is surprising, especially since STS mainly assesses the sequence-level representation quality. We investigate this by dividing the sequence pairs into two groups -- semantically similar pairs with human-annotated similarity scores no less than 0.7 and dissimilar pairs with human scores no larger than 0.3.
We plot the rank of the model inferred similarity scores against the human similarity scores in Figure \ref{fig:contragen_tok_effect} (left). As we can see, \ourmethodseq struggles in ranking semantically dissimilar sequence pairs higher and similar pairs lower. This suggests that the token-level contrastive loss is essential for making the sequence-level representations robust to spurious patterns of tokens or phrases, \eg ranking semantically similar sequences with different synonyms low and dissimilar sequences high even in presence of the same phrase (Figure \ref{fig:contragen_tok_effect} (right)).

\subsubsection{Text Generation}
\label{sec:textgen}
Next, we assess the open-ended language generation capability, where each model is required to generate text continuations given the prefixes from the WikiText-103 test set. Following \citet{simctg}, we set the lengths of prefix and continuation to 32 and 128, respectively. We use nucleus sampling \citep{nucleaus} with  $\text{top-}p = 0.95$. 
In addition to Perplexity (PPL, evaluated on the ground truth only) and MAUVE, we also evaluate the discrimination of representations of generated text under different settings in Table \ref{tab:wikitext_compare}.

\paragraph{\ourmethod Leads to More Semantically Coherent Generations} It is desired that contextual token representations within the same or semantically similar sequences have relatively higher similarities between each other when compared to similarities between tokens sampled from random contexts. Therefore,  given a prompt, lower discrimination scores are desired between the ground truth and generation, while higher discrimination values are desired between generations for randomly sampled prompts. As reported in Table \ref{tab:wikitext_compare}, compared to CLM, \ourmethod attains better semantic coherence between the ground truth and generation as indicated by Disc(S). This improvement aligns with the better MAUVE score attained by both \ourmethod and \ourmethodtok. Further, we stress that the zero valued Disc(S) attained by GPT-2 and CLM is indeed a consequence of their inferior representations since the discrimination score between semantically irrelevant sequences is also zero. Also, a slight increase in PPL is probably expected, considering that PPL is better aligned with the standard CLM objective.  Thereby, contrastive learning can be interpreted as a regularization that trades off between PPL and the desired representation properties. 




\begin{table}[t]
\centering
\resizebox{0.48\textwidth}{!}{%
\setlength{\tabcolsep}{4pt}
\begin{tabular}{@{}rcccc@{}}
\toprule
\multirow{2}{*}{Model}& \multicolumn{1}{c}{} & \multicolumn{3}{c}{Generated Text} \\  \cmidrule(l){2-2} \cmidrule(l){3-5}
&PPL$\downarrow$   
&MAUVE$\uparrow$  &Disc(S)$\downarrow$ 
&Disc(D)$\uparrow$    \\ \midrule
GPT-2        &47.50    
             &0.893   &\textbf{0.00}    &0.00   \\ 
\midrule
CLM         &\textbf{22.48}  
            &0.945    &\textbf{0.00}     &0.01   \\ 
\midrule
SimCTG      &22.51  
            &0.952  &0.11    &0.54   \\ 
\midrule
\ourmethodtok     &22.99  
            &\textbf{0.953}  &0.12     &0.49   \\ 
\hdashline\noalign{\vskip 0.5ex}
\ourmethodseq     &22.60  
            &0.933 &0.23    &\textbf{0.83}    \\ 
\hdashline\noalign{\vskip 0.5ex}
\ourmethod         &23.01  
            &0.947 &0.18     &0.62   \\ 
\bottomrule                         
\end{tabular}%
}
\vspace{-2mm}
\caption{Evaluation on the Wikitext-103 test set. Disc(D) is the discrimination score computed between the generated continuations of two randomly sampled prompts. Disc(S) is computed between the ground truth text and the generated one associated with the same prompt. A lower Disc(S) indicates better coherence with the ground truth, while a higher Disc(D) indicates better representation discrimination among the generations under different contexts. 
All metrics are evaluated over the entire test set.
}
\vspace{-2mm}
\label{tab:wikitext_compare}
\end{table}

 


%% file: sections/4.3.eval_pl.tex
\subsection{Evaluation on Programming Language}
\label{subsec:pl_eval}

In this section, we study the effectiveness of our proposed contrastive learning framework on programming language applications -- code search, code completion, and code re-ranking. Since CodeGen models are pretrained without dropout activations, we follow the same for our models in this subsection helping us study the effectiveness of \ourmethod without dropout augmentations. We also investigate how dropout would affect the decoder-only models when evaluated on the downstream tasks in Section \ref{sec:ablation} and Appendix \ref{appendix:dropout_ablation}.

\subsubsection{Code Search}
Code search is the task of retrieving relevant code fragments given a code fragment as a \emph{query}. 
We perform in-language (query and relevant code are in the same language) and cross-language (query and relevant code are in different languages) code searches.
We provide an example in Figure \ref{fig:code_to_code_search} (Appendix \ref{appendix:code_examples_and_stats}).
In this study, we experiment in the \emph{zero-shot} setting - we use the models described in Section \ref{sec:data_and_model}
to generate dense representations of code and perform a nearest neighbor search to retrieve relevant code fragments. We use publicly available implementations of \citet{guo-etal-2022-unixcoder}.\footnote{\scriptsize\url{https://github.com/microsoft/CodeBERT/tree/master/UniXcoder/downstream-tasks}}

\paragraph{Contrastive Learning Yields More Discriminative Code Representations}
For the code-to-code search task, \citet{guo-etal-2022-unixcoder} used problem solutions in Ruby, Python, and Java languages from CodeNet \citep{puri2021codenet}. 
They propose to use each program as a query and retrieve all programs that solve the same problem. We present detailed statistics of the dataset in Table \ref{tab:code_to_code_stat} (Appendix \ref{appendix:code_examples_and_stats}). 
We set the maximum sequence length as 512\footnote{We also performed experiments with maximum length 1024 but didn't observe any significant difference.} and use cosine similarity between two mean vectors of the last hidden states as relevance scores. We then sort the candidates by their scores to calculate the Mean Average Precision (MAP) score. 
We present the results for the code search tasks in Table \ref{tab:code_to_code}.\footnote{We present a comparison with encoder-only and encoder-decoder models in Table \ref{tab:code_to_code_extended} in the Appendix.}



\input{tables/main/code_to_code_main}

We observe \ourmethodtok and \ourmethod frameworks improve upon CodeGen trained with CLM by 33.5\% (absolute 2.12) and 32.6\% (absolute 2.06) on average, respectively. We also point out that the performance gap between \ourmethodtok and SimCTG are apples-to-apples comparisons since the dropout-based augmentation is not used in either models. As aforementioned, the consistently better performance of \ourmethodtok suggests the superiority of our temperature-based contrastive learning objective.
On the other hand, \ourmethodseq improves over the CLM baseline by 10.4\% only.
Code search results indicate that \ourmethodseq performs poorly compared to \ourmethodtok. This performance gap is larger than what we observed in the natural language evaluation. 
We conjecture that \ourmethodtok generates better discriminative representations for code sequences since the finer-grained understanding of the code tokens is crucial to understanding the code sequences' functionality (semantics).
To verify this, we check if non-semantic factors impact model performances in the following section. 

\begin{figure*}
\centering
\begin{subfigure}[b]{0.9\textwidth}
   \includegraphics[width=1\linewidth]{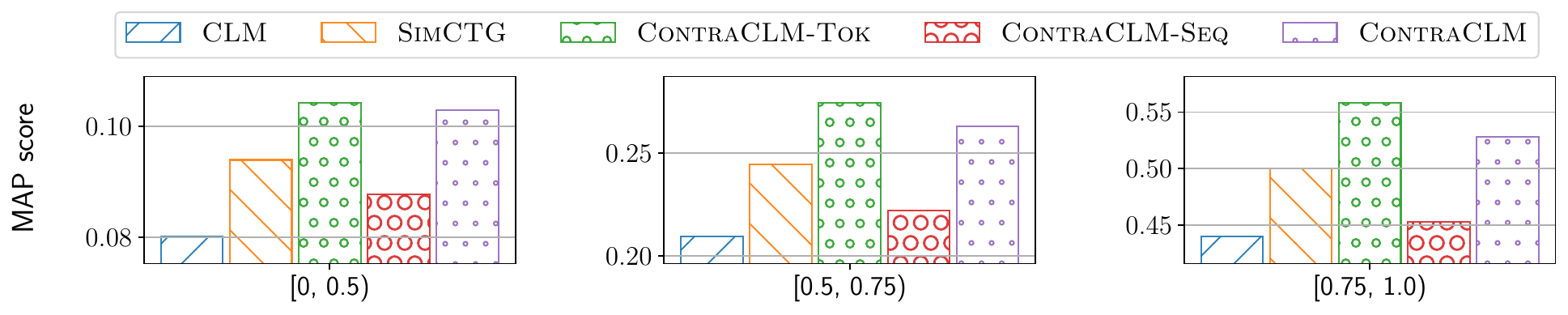}
   \caption{Performance breakdown based on edit similarities
(x-axis).}
   \label{fig:es_vs_perf_py} 
\end{subfigure}

\medskip
\begin{subfigure}[b]{0.9\textwidth}
   \includegraphics[width=1\linewidth]{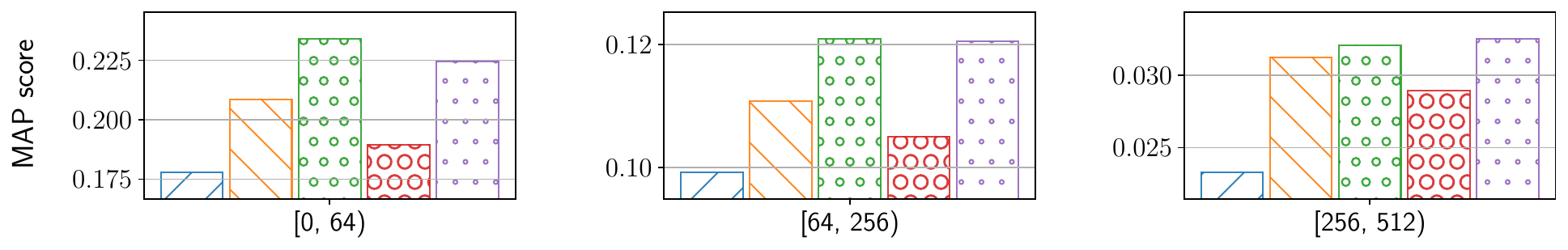}
   \caption{Performance breakdown based on length differences (x-axis).}
   \label{fig:len_vs_perf_py}
\end{subfigure}
\caption{Code search performances based on (a) and (b) between the query code fragments (in Python) and their relevant code fragments (in Python). We observe that in both cases, \ourmethodtok outperforms CLM, SimCTG, and \ourmethodseq.}
\vspace{-2mm}
\label{fig:es_len_vs_perf_py}
\end{figure*}



\paragraph{Token-level Contrastive Learning is Effective for Code Understanding}
We break down the code search performance based on edit similarities and length differences between query code and their relevant code fragments.
While edit similarity indicates how much queries and their relevant code overlap, the length difference indicates whether models effectively capture relevance between two code fragments if they are similar in length or differ significantly.
We present the results for Python language in Figure \ref{fig:es_len_vs_perf_py} (for all the languages, see Figures \ref{fig:es_vs_perf_code_search} \& \ref{fig:length_vs_perf_code_search} in Appendix \ref{appendix:pl_details}).
The results show that \ourmethodtok outperforms CLM, SimCTG, and \ourmethodseq irrespective of edit similarities and length differences. 
Therefore, we can conclude that sequence overlap or length are not the reasons for improvements in \ourmethodtok. Presumably, a finer-grained understanding of code tokens makes \ourmethodtok more effective for code representations.




\subsubsection{Code Completion and Re-Ranking}
\label{sec:code_completion}


Given a sequence of tokens composed of natural language, function signature, and input-output examples (as a whole, we call them prompt), the goal of the code completion task is to complete the function. To evaluate the functional correctness of a complete code, we use existing benchmarks that include unit tests. If the generated code successfully passes the unit tests, we refer to this as successful execution. We compute pass@$k$ for $k \leq n$ following \cite{codex}.
In addition, we compare the models on the code re-ranking task -- given $n$ sampled code using a code completion model, the goal is to order the generated samples, for which we use the mean log probability of each sampled code~\citep{codex}. For code re-ranking evaluation, we report ranked pass@$k$ \citep{inala2022fault}. Figure \ref{fig:humaneval_ex} (Appendix \ref{appendix:code_examples_and_stats}) illustrates both the code completion and re-ranking tasks. We detail the evaluation metrics in Appendix \ref{appendix:pl_eval_metrics}.

\input{tables/main/humaneval_main}

\paragraph{Contrastive Learning Improves Source Code Generation}
\citet{codex} introduced HumanEval, a collection of 164 handwritten programming problems and their respective unit tests. Each problem in this dataset is presented using a prompt for a function, and the task is to complete the function, such that it can pass all unit tests.
In all our experiments, we use nucleus sampling \citep{nucleaus} with top $p = 0.95$.
We sample $n = 10$ completions per problem with sampling temperature $0.2$.
Table \ref{tab:humaneval} presents the evaluation results on the HumanEval benchmark.

While \ourmethodtok and \ourmethodseq perform comparably to CLM and SimCTG, 
\ourmethod outperforms them significantly, \ie by 9\% and 10.3\% in terms of pass@1 accuracy respectively, and
by 11\% and 12\% in terms of ranked pass@1 accuracy, respectively.
While \ourmethodseq underperforms in code completion, it boosts code re-ranking significantly. We hypothesize the improvement is due to the contrastive learning's alignment with the mean log probability-based re-ranking choice.

%% file: tables/main/code_to_code_main.tex
\begin{table*}[t]
    \begin{center}
    {\normalsize{
        \resizebox{0.9\linewidth}{!} {%
        \def\arraystretch{1.0}%
        \begin{tabular}{r c c c  c c c c c c c}
        \toprule
        \multirow{2}{*}{Model} & \multicolumn{3}{c}{Ruby}  & \multicolumn{3}{c}{Python} & \multicolumn{3}{c}{Java} & \multirow{2}{*}{Avg.} \\ 
        \cmidrule(lr){2-4} 
        \cmidrule(lr){5-7}
        \cmidrule(lr){8-10}
        & Ruby & Python & Java & Ruby & Python & Java & Ruby & Python & Java & \\ 
        \midrule
        CodeGen         & 16.18 & 5.90 & 0.52 & 2.66 & 18.11 & 0.36 & 1.61 & 1.65 & 10.16 & 6.35 \\
        \midrule
        CLM             & 16.36 & 6.67 & 0.80 & 3.07 & 15.72 & 0.46 & 1.41 & 2.11 & 10.25 & 6.32 \\
        \midrule
        SimCTG          & 17.66	& 7.19 & 1.94 & 7.63 & 18.31 & 1.78 & 1.63 & 2.32 & 10.83 & 7.70 \\
        \midrule
        \ourmethodtok         & \textbf{18.02} & \textbf{7.84} & 2.51 & 8.76 & \textbf{20.46} & 2.48 & \textbf{1.91} & \textbf{2.58} & \textbf{11.43} & \textbf{8.44} \\
        \hdashline\noalign{\vskip 0.5ex}
        \ourmethodseq         & 16.76	& 5.45 & 1.06 & 7.40 & 16.74 & 1.41 & 1.55 & 2.25 & 10.23 & 6.98 \\
        \hdashline\noalign{\vskip 0.5ex}
        \ourmethod             & 17.90 & 7.78 & \textbf{2.56} & \textbf{9.05} & 19.74 & \textbf{2.64} & 1.90 & 2.50 & 11.32 & 8.38 \\
        \bottomrule
        \end{tabular}
        }
    }
    }
    \vspace{-2mm}
    \caption{
    MAP score (\%) of the zero-shot code search task. The language names mentioned in the top two rows indicate the languages queries and candidates are written in.
    }
    \vspace{-2mm}
    \label{tab:code_to_code}
    \end{center}
\end{table*}

%% file: tables/main/humaneval_main.tex
\begin{table}
\centering
\setlength{\tabcolsep}{4pt}
\def\arraystretch{1.0}%
\resizebox{\linewidth}{!} {%
\begin{tabular}{r c c c c}
    \toprule
    \multirow{2}{*}{Model} & \multicolumn{2}{c}{Pass@k}  & \multicolumn{2}{c}{Ranked Pass@k} \\
    \cmidrule(lr){2-3} 
    \cmidrule(lr){4-5}
    & k=1 & k=5 & k=1 & k=5 \\ 
    \midrule
    CodeGen     & 12.65 & 16.89 & 13.42 \textsubscript{(+0.77)} & 17.07 \textsubscript{(+0.18)} \\ 
    \midrule
    CLM         & 13.42 & 18.08 & 15.38 \textsubscript{(+1.96)} & 18.29 \textsubscript{(+0.21)} \\
    \midrule
    SimCTG      & 13.26 & 17.29 & 15.24 \textsubscript{(+1.98)} & 18.29 \textsubscript{\textbf{(+1.00)}} \\
    \midrule
    \ourmethodtok     & 12.96 & 17.01 & 15.24 \textsubscript{(+2.96)} & 17.68 \textsubscript{(+0.67)} \\
    \hdashline\noalign{\vskip 0.5ex}
    \ourmethodseq     & 13.64 & 15.85 & 16.99 \textsubscript{\textbf{(+3.35)}} & 16.46 \textsubscript{(+0.61)} \\
    \hdashline\noalign{\vskip 0.5ex}
    \ourmethod         & \textbf{14.63} & \textbf{18.83} & \textbf{17.07} \textsubscript{(+2.44)} & \textbf{18.90} \textsubscript{(+0.07)} \\
    \bottomrule
\end{tabular}
}
\vspace{-2mm}
\caption{
Evaluation results on the HumanEval benchmark. The numbers in the subscript indicate the difference between ranked pass@k and pass@k. While \ourmethodtok and \ourmethodseq perform competitively to the baselines, \ourmethod significantly outperforms them.
}
\vspace{-2mm}
\label{tab:humaneval}
\end{table}

%% file: sections/4.4.ablation.tex
\subsection{Discussion}
\label{sec:ablation}

\paragraph{Impact of Dropout} 
Dropout-based augmentation \citep{gao2021simcse} for contrastive learning on language models has shown to have a significant improvement on discriminative tasks. We observe the same trend on both GPT-2 and CodeGen (see Table \ref{tab:discriminative_tasks_dropout} in Appendix \ref{appendix:dropout_ablation}). However, we observed the opposite for language generation, no matter when training with CLM only or with contrastive learning (see Table \ref{tab:generative_tasks_dropout} in Appendix \ref{appendix:dropout_ablation}). Dropout has been one of the key ingredients for training large models. Further investigation on proper ways to use and evaluate it is indeed required. Nevertheless, even without dropout, Section \ref{subsec:pl_eval} shows \ourmethod still yields significant improvement. 

\paragraph{Bridge the Gap}
In comparison with the causal (left-to-right) attention mechanism of the decoder-only models, the bidirectional attention mechanism better leverages the context of sequences, yielding better representations for discriminative tasks. Take the encoder-only models as an example: as Table \ref{tab:STS_encoder_decoder} in Appendix shows, both BERT-Base \citep{devlin2018bert} and RoBERTa-Base \citep{liu2019roberta} outperform GPT-2 by at least 60\% relative performance on STS. Although the performance gap between CodeGen and the encoder-only or encoder-decoder models decreases in Table \ref{tab:code_to_code_extended}, it is still significant considering that both the model and pretraining data sizes used by CodeGen are much larger. Such a large performance gap severely limits the usage of decoder-only models in many discriminative tasks. On the bright side, contrastive learning shows the promise to bridge the gap, \eg reducing the relative performance gap between GPT-2 and the encoder-only models by at least $50\%$ when evaluating on STS (see Table \ref{tab:STS_encoder_decoder}).  Please refer to Appendix \ref{appendix:bridge_the_gap} for more detailed discussions. 

%% file: sections/5.conclusion.tex
\section{Conclusion}
In this paper, we present \ourmethod, an effective contrastive learning framework to resolve the representation degeneration issue of CLMs trained with the autoregression objective.  We assess the effectiveness of \ourmethod on various downstream tasks in both the natural language and code domains, where we attain significant improvements on both discrimination and generation tasks. 
While we explored only the decoder-only CLMs, our proposed contrastive learning framework can serve as a drop-in term for encoder-decoder, encoder-only, or prefixLM models also. We leave these explorations as future work.


\section*{Limitations}
While our work displays many strengths, we highlight some limitations. First, we focus on Python for programming language evaluation, which is one of the widely used programming languages. However, we believe our proposed approach \ourmethod would benefit Code LMs with any language support.
Second, the empirical findings this work presents are mainly based on the smaller version of GPT-2 and CodeGen, with 124M and 350M parameters, respectively. However, as shown in Figure \ref{fig:effective_contraclm}, by continuing to train the pretrained models with our proposed objective, \ourmethod is able to address not only the isotropy and poor discrimination issue that both GPT2-small and CodeGen suffer from but also improve the representation quality of GPT2-large which has a good starting point for both isotropy and discrimination. Therefore, we believe the effectiveness of \ourmethod should be applicable to larger versions of these LMs, regardless of whether they suffer from the anisotropy issue (\eg large CodeGen models) or not (large scale GPT-2 models). We leave the explorations of larger models as future work. 

\section*{Ethics Statement}
\paragraph{Training data} We use WikiText-103 and source code in Python from permissively licensed GitHub repositories to train GPT2 and CodeGen, respectively.
We do not perform any preprocessing that would get rid of any personally identifiable information or offensive content. However, the use of code LMs comes with certain risks, e.g., generating biased, toxic, and insecure code. We refer readers to \citet{codex} (Section 7) for a detailed discussion on the broader impact of code LMs.

\paragraph{Compute} We use an in-house cluster of 128 A100s for all jobs in this paper. Each run takes a couple of hours to a day, depending on the configuration and the model size. We performed one round of training for each setting as it is very expensive to repeat them multiple times. However, we perform the code completion and re-ranking evaluation with three seeds. STS and code search evaluation do not need multiple runs of inference (as the predictions are deterministic).

%% file: appendix.tex
\twocolumn[{%
 \centering
 \Large\bf Supplementary Material of Paper 4482: Appendices \\ [20pt]
}]

\section{Contrastive Learning for CLM}
\label{appendix:model}
We detail our proposed token-level and sequence-level contrastive losses. Before that, we first call out the following notations that will be used throughout this section. Let $\mathbf{x} = [\mathbf{x}_1, \mathbf{x}_2, \cdots, \mathbf{x}_{|\mathbf{x}|}]$ denote a sequence with variable length $|\mathbf{x}|$, \eg a text document or a code snippet, and $\mathbf{h} = [\mathbf{h}_1, \mathbf{h}_2, \cdots, \mathbf{h}_{|\mathbf{x}|}]$ be its representation output by the last hidden layer of the decoder. For a randomly sampled batch $\mathcal{B} = \left\{\mathbf{x}^j\right\}_{j=1}^{N}$ with $N$ sequences, we use $\mathbf{x}_i^j$ and $\mathbf{h}_i^j$ to denote the $i^{th}$ token and its representations in the $j^{th}$ sequence, respectively. Let $\mathbf{h}^j, \mathbf{h}^{j^+}$ denote the representation pair of sequence $\mathbf{x}^j$ and $\mathbf{h}^j_i, \mathbf{h}^{j+}_{i}$ correspond to the representations of the $i$-th token. Such representation pairs are referred to as positive pairs in contrastive learning,  which are often obtained via data augmentation.

\input{tables/appendix/equations}

\subsection{Token-Level Contrastive Learning}
As aforementioned, $\mathbf{h}^j_i, \mathbf{h}^{j^+}_{i}$  are a pair of representations  for $\mathbf{x}^j_i$, the $i$-th token in the $j$-th sequence. Let $\mathcal{I}_j = \left\{1, 2, \dots, |\mathbf{x}_j|\right\}$ denote the indices of tokens in $\mathbf{x}_j$. Further let $\tau$ denote the temperature hyper-parameter and $\diamond$ denotes the cosine similarity, \ie $\mathbf{a} \diamond \mathbf{b} = \mathbf{a}^T \mathbf{b} / \|\mathbf{a}\|_2 \|\mathbf{b}\|_2$.  Then we minimize $\mathcal{L}_{\text{Tok}}$ defined in Table~\ref{tab:equations}.


\subsection{Sequence-Level Contrastive Learning}
Let $\mathcal{I}_B = \{1, 2, \dots, N\} \cup \{1^+, 2^+, \dots, N^+\}$ denote indices of all $2N$ sequence-level representations for batch $\mathcal{B}$. The sequence-level contrastive loss is defined as $\mathcal{L}_{\text{Seq}}$ in Table~\ref{tab:equations}.


\section{Training Details}
\label{appendix:training}

\paragraph{Training Data} For text, we use WikiText-103, a collection of over 100 million tokens extracted from the set of verified and featured articles on Wikipedia \citep{merity2016pointer}. For code, we collect permissively licensed Python code from GitHub. Following \cite{codex,codegen}, we perform filtering and deduplication and further remove data that contains a significant use of non-English languages or is not parsable, resulting in a dataset of 101GB code.

\paragraph{Model} We use GPT-2 \citep{gpt2} and CodeGen 350M monolingual \citep{codegen} for all experiments on natural language (text) and  programming language (code), respectively.  We set the batch size to 512 and continue to train GPT-2 on WikiText-103 and CodeGen on the GitHub data for 12 and 2 epochs, respectively. We train both models using a max sequence length of 512 tokens and 1024 for WikiText-103 and Code data, respectively. We set the learning rate to 2e-5, warm-up steps as 500 with linear annealing after peak learning rate, weight decay of 0.1, temperature of 0.05 (when using contrastive losses), and gradient clipping of 1.0. We use AdamW optimizer \citep{loshchilov2018decoupled} with $\beta_1=0.9$, $\beta_2=0.999$, and $\epsilon=10^{-8}$ following \citep{codegen}. Our training pipeline is based on PyTorch Lightning\footnote{\scriptsize\url{https://www.pytorchlightning.ai/}}, and we use DeepSpeed \citep{deepspeed} for training optimization.

\paragraph{Processing Code Training Data}
Our preprocessing strategy for code datasets used for training is designed to ensure that we optimize for data utilization while retaining the syntactic structure of programming language sequences. We also eliminate duplicate sequences since this benefits training large language models~\citep{Lee2022DeduplicatingTD}. Specifically, we break long sequences into chunked sequences of smaller lengths to retain most parts of the original program. Further, we maintain syntactic structure in the chunks by ensuring that each chunk ends with a `\texttt{\textbackslash n}' character.
Each chunk obtained this way contains at most \texttt{max\_chars\_per\_seq} characters where \texttt{max\_chars\_per\_seq = max\_tokens\_per\_seq * chars\_per\_tok}. In our experiments, we fix \texttt{chars\_per\_tok}~$=~3.2$ and  \texttt{max\_tokens\_per\_seq}~$= 1024$. We also perform deduplication using character-based exact matches between chunked sequences over the entire dataset. This step helps eliminate exact duplicates that might be present after the chunking stage.

\begin{figure*}[!ht]
     \centering
     \begin{subfigure}[b]{0.9\textwidth}
         \centering
         \includegraphics[width=\textwidth]{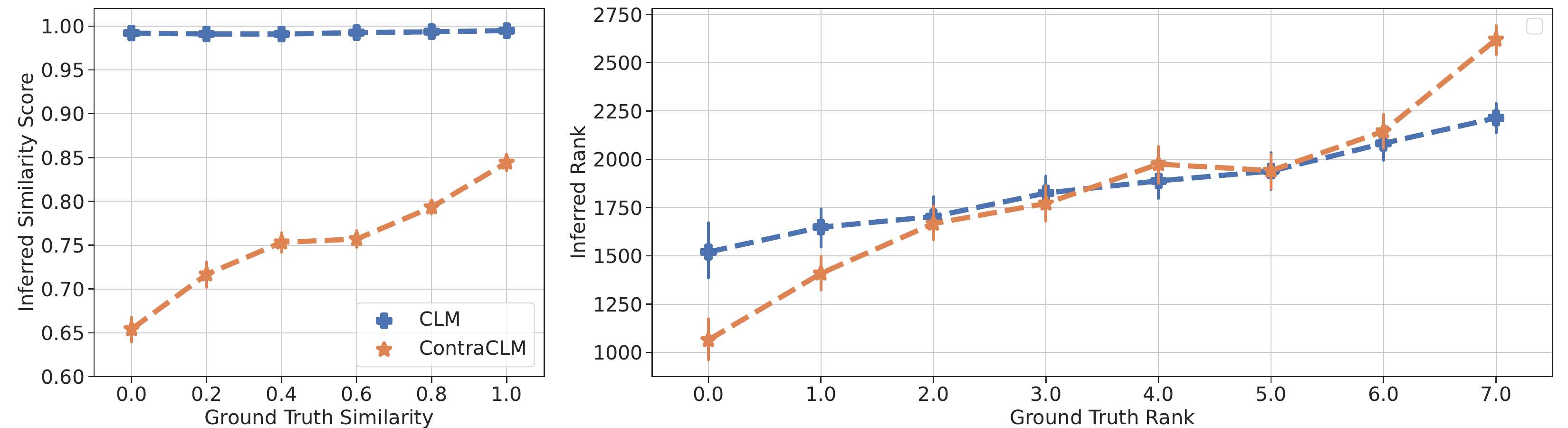}
         \caption{STS14: \textbf{(Left)} Predicted cosine similarity vs. human annotated ground truth. \textbf{Right} Similarity ranking according to the model predicted similarity scores vs Human similarity based ranking.}
         \label{fig:appendix_sts_scores_full}
     \end{subfigure}
     
     \bigskip
     \begin{subfigure}[b]{0.9\textwidth}
         \centering
         \includegraphics[width=\textwidth]{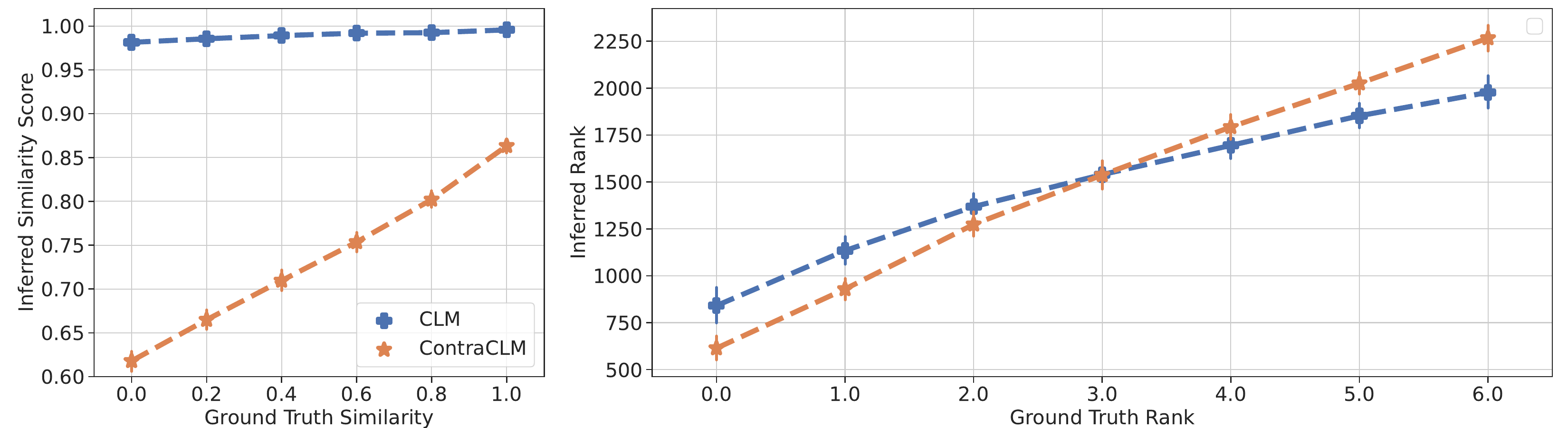}
         \caption{STS15: \textbf{(Left)} Predicted cosine similarity vs. human annotated ground truth. \textbf{Right} Similarity ranking according to the model predicted similarity scores vs Human similarity-based ranking. }
         \label{fig:appendix_sts_rank_full}
     \end{subfigure}
        \caption{CLM versus Contrastive Learning in Similarity Prediction and Ranking. We report the results on two STS benchmarks: (1) \textit{STS14} where CLM performs the worst compared to its performance on the other STS tasks; and \textit{STS15} where CLM attains the best performance when comparing with its own. For the purpose of illustration, we scale the human-annotated similarity scores from $[0, 5]$ to $[0, 1]$. A good CLM is expected to predict discriminative similarity scores such that the resulting ranking results are as close to the ranks provided by humans as possible.}
        \vspace{-2mm}
        \label{fig:appendix_sts_rank_and_score_analysis}
\end{figure*}

\section{More on Evaluation}

\subsection{Representation Quality Evaluated on STS}
\label{appendix:sts_analysis}
For each sequence pair in STS, a fine-grained similarity score ranging from 0 to 5 is provided, with a high similarity score indicating semantically similar pairs and low similarity scores suggesting semantically dissimilar or irrelevant pairs. For better illustration, we scale the human-annotated similarity scores to $[0, 1]$ to align with the model-predicted cosine similarity scores. This does not affect the evaluation as the spearman correlation reported in Section \ref{sec:nl_experiment} is a rank-based correlation metric.   

\paragraph{CLM yields poorly discriminative representations}{We report the model predicted similarity scores of sequence pairs in the left column in Figure \ref{fig:appendix_sts_rank_and_score_analysis}. A good model is expected to yield representations that attain higher similar scores between similar sequence pairs and lower similarity values for dissimilar sequences. Thereby, a large gap between the predicted similarity scores of similar and dissimilar pairs is desired. However, as seen in Figure \ref{fig:appendix_sts_rank_and_score_analysis} (left), the similarity scores attained by the model trained with the standard CLM only objective are almost at one for both similar and dissimilar sequence pairs. This suggests that the representations yielded by CLM can\footnote{As investigated in Figure \ref{fig:motivation}, the decoder-only models pretrained with the CLM-only can suffer from the anisotropy issue, which depends on the model size and domain.} be squeezed into a tiny cone in the representation space rather than being scattered apart to leverage the vector space's capacity better. 
Despite the resulting similarity ranks are not entirely flattened, as shown in the right column in Figure \ref{fig:appendix_sts_rank_full}, CLM struggles in ranking similar sequences lower and dissimilar sequences higher as a consequence of the poorly discriminative representations such that low similarity scores can be assigned to the semantically similar and high similarity scores to dissimilar sequence pairs.

In contrast, Figure \ref{fig:appendix_sts_rank_and_score_analysis} (left) further validates that contrastive learning effectively yields more discriminative representations with a comparatively larger similarity gap between similar pairs and dissimilar pairs. Thereby, the similarity ranking results of the sequence pairs are more aligned with those obtained according to similarity scores provided by a human, as shown in Figure \ref{fig:appendix_sts_rank_and_score_analysis} (right). 
}

\subsection{Programming Language Evaluation}

\subsubsection{Evaluation Metrics}
\label{appendix:pl_eval_metrics}

\paragraph{Mean Average Precision (MAP)} For a set of queries, it indicates the mean of the average precision scores for each query.
\begin{equation*}
    MAP = \frac{\sum_{q=1}^Q AveP(q)}{Q}
\end{equation*}
where $Q$ is the number of queries.

\paragraph{Pass@k} Given a problem (code prompt as shown in Figure \ref{fig:humaneval_ex}), pass@k indicates the functional correctness of model-generated code samples. A problem is considered solved if any sample passes the unit tests. Following \cite{codex}, we generate $n \geq k$ samples per problem (in this paper, we use $n=10$ and $k \in \{1, 5\}$), count the number of correct samples $c \leq n$ that pass unit tests, and calculate the unbiased estimator of pass@$k$ as:
\begin{equation*}
    \text {pass@}k := \underset{Problems}{\mathbb{E}}\left[ 1 - \frac{{{n-c} \choose k}}{{n \choose k}} \right] \;.
\end{equation*}

\paragraph{Ranked Pass@k}
Unlike Pass@k, where we randomly chose $k$ out of $n$ samples, in ranked pass@k, we chose the top-$k$ samples based on model-provided scores and then computed pass@k. 

\input{tables/appendix/code_to_code_stat}

\subsubsection{Examples and Statistics}
\label{appendix:code_examples_and_stats}
In Figure \ref{fig:code_to_code_search}, we present an example of a query code fragment in Python and relevant code fragments in Python and Java, respectively. While in-language code-to-code search refers to retrieving relevant code fragments in the same language, cross-language code-to-code search refers to retrieving code fragments in a different language. We present the statistics of the code search dataset in Table \ref{tab:code_to_code_stat}.
To demonstrate the code completion task, we illustrate an example in Figure \ref{fig:humaneval_ex}.

\subsubsection{Detailed Code Search Results}
\label{appendix:pl_details}

We provide a comparison between encoder-only \citep{codebert, graphcodebert}, encoder-decoder \citep{plbart, codet5}, and decoder-only models (main focus of this work) on the zero-shot code-to-code search task in Table \ref{tab:code_to_code_extended}. We see that \ourmethodtok and \ourmethod outperform the encoder-only model CodeBERT and both the encoder-decoder models.
It is important to note that the comparison across these models is not apple-to-apple as these models differ in size, the scale of pretraining, and language settings. This comparison's purpose is to show the promise of decoder-only models being used in discriminative tasks like code search.

\begin{figure*}[ht]
    \centering
    \includegraphics[width=0.75\textwidth]{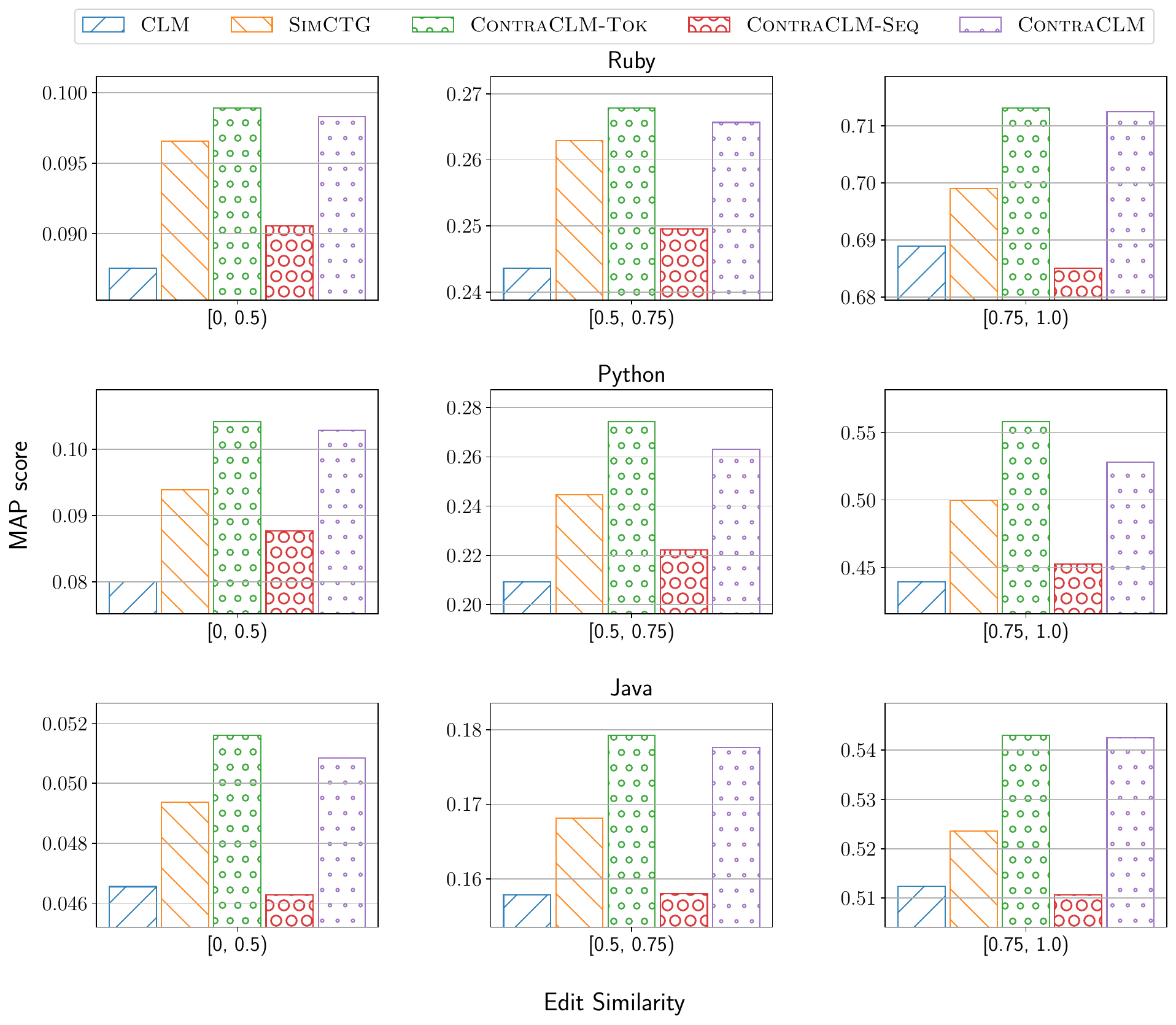}
    \caption{Code-to-code search performance (MAP score) breakdown for different models based on the edit similarities ([0, 1]) between the query code fragments and the relevant code fragments. Higher and lower edit similarity indicates the search task is trivial or difficult, respectively.}
    \label{fig:es_vs_perf_code_search}
\end{figure*}

\begin{figure*}[ht]
    \centering
    \includegraphics[width=0.75\textwidth]{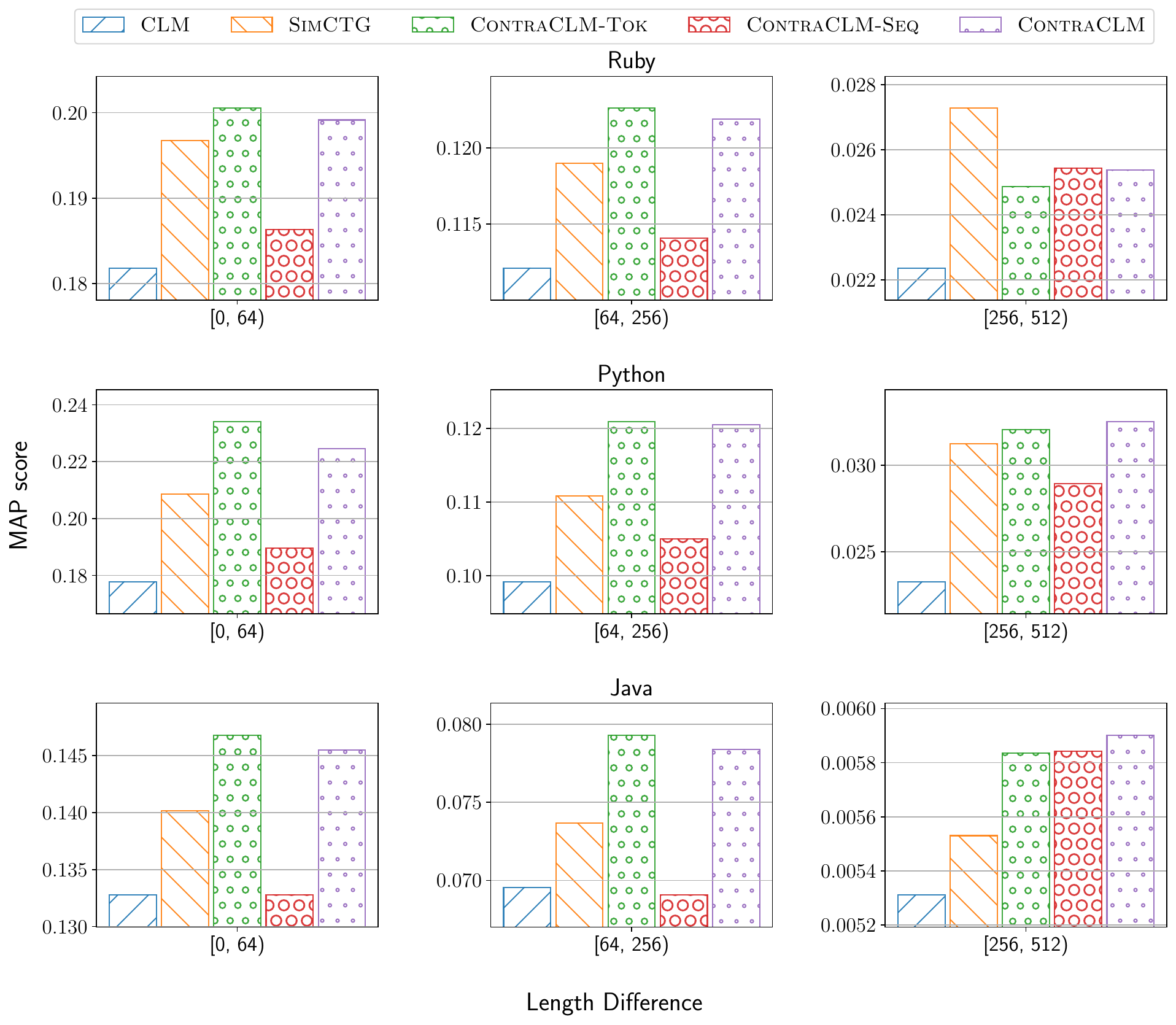}
    \caption{Code-to-code search performance (MAP score) breakdown based on (absolute) length differences between the query code fragments and their relevant code fragments.}
    \label{fig:length_vs_perf_code_search}
\end{figure*}

We further break down the code search performances based on edit similarities and length differences between query code and their relevant code fragments. We present the results in Figure \ref{fig:es_vs_perf_code_search} and \ref{fig:length_vs_perf_code_search}.
We observe a similar performance trend in all three languages, although cross-lingual search performance still needs to improve.
Nonetheless, the objective of this performance analysis is to show that sequence overlap or length are not the reasons for improvements in \ourmethodtok. Instead, a finer-grained understanding of code tokens due to the token-level contrastive learning makes \ourmethodtok more effective.

\input{tables/appendix/code_to_code_extended}

\section{More Analysis and Discussions}

\subsection{Bridge the Gap on Discriminative Tasks}
\label{appendix:bridge_the_gap}
Compared to the causal (left-to-right) attention mechanism of the decoder-only models, the bidirectional attention mechanism in both encoder-only and encoder-decoder models allows for better leverage of the context of the sequence and hence leads to better representations. 

Taking the encoder-only models in Table \ref{tab:STS_encoder_decoder} for illustration, on average, BERT-Base \citep{devlin2018bert} and Roberta-Base \citep{liu2019roberta} outperform GPT-2 with 67.25\% (absolute 21.17\%) and 84.62\% (absolute 26.64\%) relative improvement on STS, respectively. Although the performance gap between CodeGen and the BERT models trained on programming languages, \ie CodeBERT \citep{codebert} and GraphCodeBERT \citep{graphcodebert}, decreases or even diminishes when evaluated on the code search tasks, the performance gap is still significant as both the model size and pretraining data in CodeGen are much larger than those used by the encoder-only models in Table \ref{tab:code_to_code_extended}. 
Similar trends were observed in the performance gap between the decoder-only and encoder-decoder models on both natural language \citep{textbart, textt5} and programming language \citep{plbart, codet5}.

\input{tables/appendix/code_to_code_dropout}

\input{tables/appendix/human_eval_dropout}

\begin{table*}[htbp]
    \begin{center}
    \resizebox{0.99\textwidth}{!}{
    \begin{tabular}{rcccccccc}
   
    \toprule
    \text{Model}& \text{STS12} & \text{STS13} & \text{STS14} & \text{STS15} & \text{STS16} & \text{SICK-R} & \text{STS-B} & \text{Avg.} \\ 
     \cmidrule(l){1-9} 
    SimCTG &  30.32&	37.10&	31.99&	39.68&	42.73&	46.26&	25.27&	36.19 \\
    \ourmethod$_{- \text{Dropout}}$& 38.22	&40.15	&33.57	&53.16	&45.35	&47.47	&36.10	&42.00    \\
    \cmidrule(l){1-9} 
     SimCTG$_{+ \mathcal{L}_\text{Seq} + \text{Dropout}}$ &\textbf{38.70}	 &43.60	&36.29	&50.01	&45.19	&48.25	&33.36	&42.20 \\
    \ourmethod$_{+ \text{Dropout}}$ & 37.54	  &\textbf{45.23}	&\textbf{36.41}	 &\textbf{56.74}   &\textbf{50.30}   &\textbf{51.52}    &\textbf{39.49}   &\textbf{45.32} \\
    \bottomrule
    \end{tabular}
}
    \caption{\ourmethod outperform SimCTG \citep{simctg} even without dropout-based data augmentation (first two rows); or augmenting SimCTG with dropout and sequence-level contrastive loss defined in Table \ref{tab:equations}.}
    \label{tab:contragen_vs_simctg}
  \end{center}
\end{table*}

The large performance gap severely limits the decoder-only models used in many discriminative tasks. To this end, contrastive learning shows the promise to largely bridge the gap. As seen in Table \ref{tab:STS_encoder_decoder}, on STS, \ourmethod reduces the relative performance gap from 67.24\% (absolute 21.12\%) to  16.17\% (absolute 7.33\%) regarding BERT-Base, and from 84.62\% (absolute 26.64\%) to 28.24\% (absolute 12.8\%). 
Similarly, Table \ref{tab:code_to_code_extended} shows that \ourmethod outperforms encoder-decoder models and performs comparably to the encoder-only model, GraphCodeBERT.

\subsection{Dropout for Contrastive Learning}
\label{appendix:dropout_ablation}

\citet{gao2021simcse} showed that the dropout-based augmentation is an effective strategy for unsupervised contrastive learning, and the follow-up works \citep{diffcse, wu2021esimcse} endorse the effectiveness. This motivates us to study dropout-based augmentation in our proposed contrastive learning framework. We present the results on discriminative and generation tasks in Tables \ref{tab:discriminative_tasks_dropout} and \ref{tab:generative_tasks_dropout}, respectively.
From the results, it is evident that the adoption of dropout-based augmentation improves the discrimination task performances, which corroborates the findings of \cite{gao2021simcse}. In contrast, dropout-based augmentation hurts the generation task performances. 
While for code completion, we anticipated that dropout-based augmentation would hurt the performances since we use CodeGen \citep{codegen} which does not use dropout during the original pretraining stage.
However, we observe a drop in perplexity due to disabling the dropout for both CLM and \ourmethod in Table \ref{tab:generative_tasks_dropout}, which does not go with our anticipation, especially considering that, unlike CodeGen, GPT-2 is pretrained with dropout enabled. We leave this as a future exploration to dive deeper into the reasoning behind this finding.

\subsection{\ourmethod outperforms SimCTG}
\label{appendix:simctg_vs_contragen}
To better understand the performance gap between \ourmethod and SimCTG \citep{simctg}, we run the following ablations on GPT-2 and report the evaluations on STS. In Table \ref{tab:contragen_vs_simctg}, we report the results of (1) running \ourmethod w/o dropout-based data augmentation and compare it with the original SimCTG model and (2) augmenting SimCTG with both the sequence-level contrastive loss and dropout-based augmentation and compare it with our proposed \ourmethod model. As we can see, \ourmethod consistently outperforms SimCTG in both settings. Figure \ref{tab:contragen_vs_simctg} together with our results reported in Section \ref{subsec:pl_eval} where we disabled the dropout-based augmentation for \ourmethod and its variations, but still observed consistent better performance than SimCTG on both discrimination and generation tasks, conclude that \ourmethod is better than SimCTG across domains and settings.

\input{figures/code_search_ex}

\input{figures/humaneval}

%% file: tables/appendix/equations.tex
\begin{table*}[!htbp]
\renewcommand{\arraystretch}{1.3}
\centering
\begin{tabular}{|c|c|}
\hline 
Contrastive Loss & Expression \\[0.6ex] 
\hline 
$\displaystyle\mathcal{L}_{\text{Tok}}$ & {\small$\begin{aligned}\displaystyle \sum_{j=1}^{N} \sum_{i=1}^{|\mathbf{x}^j|} - \Biggl(\log \frac{\exp(\mathbf{h}_i^j \diamond \mathbf{h}^{j^+}_{i}/\tau)}{\exp(\mathbf{h}_i^j \diamond \mathbf{h}^{j^+}_{i}/\tau) + \sum_{t \in \mathcal{I}_j \setminus i} \left[\exp(\mathbf{h}^j_i \diamond \mathbf{h}^j_{t}/\tau) + \exp(\mathbf{h}^j_i \diamond \mathbf{h}^{j^+}_{t}/\tau)\right]}  \\ + \log \frac{\exp(\mathbf{h}^{j^+}_{i} \diamond \mathbf{h}_i^j /\tau)}{\exp(\mathbf{h}^{j^+}_{i} \diamond \mathbf{h}_i^j /\tau) + \sum_{t \in \mathcal{I}_j \setminus i} \left[\exp(\mathbf{h}^{j^+}_{i} \diamond \mathbf{h}^j_{t}/\tau) + \exp(\mathbf{h}^{j^+}_{i} \diamond \mathbf{h}^{j^+}_{t}/\tau)\right]} \Biggl) \end{aligned}$}\\ \hline
$\displaystyle \mathcal{L}_{\text{Seq}}$ & {\small$\begin{aligned}\displaystyle  \sum_{j = 1}^{N}  - \Biggl(\log \frac{\exp(\mathbf{h}^j \diamond \mathbf{h}^{j^+}/\tau)}{\exp(\mathbf{h}^j \diamond \mathbf{h}^{j^+} /\tau) + \sum_{k \in \mathcal{I}_{\mathcal{B}} \setminus j, j^+ } \exp(\mathbf{h}^j \diamond \mathbf{h}^{k}/\tau)}  \\ + \log \frac{\exp(\mathbf{h}^{j^+} \diamond \mathbf{h}^{j}/\tau)}{\exp(\mathbf{h}^{j^+} \diamond \mathbf{h}^{j}/\tau) + \sum_{k \in \mathcal{I}_{\mathcal{B}} \setminus j, j^+ } \exp(\mathbf{h}^{j^+} \diamond \mathbf{h}^{k}/\tau)} \Biggr)\end{aligned}$} \\ 
\hline
\end{tabular}
\caption{Formulation of our token-level and sequence-level contrastive losses denoted as $\mathcal{L}_{\text{Tok}}$ and $\mathcal{L}_{\text{Seq}}$ respectively.}
\label{tab:equations}
\end{table*}

%% file: tables/appendix/code_to_code_stat.tex
\begin{table*}[htbp]
    \begin{center}
    {\normalsize{
        \def\arraystretch{1.0}%
        \begin{tabular}{l l l l}
        \toprule
        & Ruby  & Python & Java \\ 
        \midrule
        Total problems                  & 1,708 & 2072 & 3142 \\
        Total \#solution                & 11,744 & 15,594 & 23,530 \\
        Avg. \#solution / problem       & 6.9 & 7.5 & 7.5 \\
        \midrule
        Avg. length / solution         & 160.4 & 214.4 & 894.9  \\
        Stdev. of length / solution (problem-wise) & 112.80 & 113.79 & 813.0  \\
        Solutions with length $> 512$    & 409 & 1,200 & 10,023 \\
        Solutions with length $> 1024$   & 78 & 278 & 4,766 \\
        \midrule
        Avg. edit similarity            & 0.48\textsubscript{(+0.13)} & 0.52\textsubscript{(+0.13)} & 0.49\textsubscript{(+0.13)} \\
        \bottomrule
        \end{tabular}
        }
    }
    \vspace{-2mm}
    \caption{
    Statistics of code-to-code search task dataset created from CodeNet \citep{puri2021codenet}. We truncate the code if its length exceeds the maximum sequence length, which is set to 512. 
    }
    \vspace{-2mm}
    \label{tab:code_to_code_stat}
    \end{center}
\end{table*}

%% file: tables/appendix/code_to_code_extended.tex
\begin{table*}[htbp]
\begin{subtable}{1\textwidth}
    \resizebox{\linewidth}{!}{%
    \centering
    \begin{tabular}{rcccccccc}
   
    \toprule
    \text{Model}& \text{STS12} & \text{STS13} & \text{STS14} & \text{STS15} & \text{STS16} & \text{SICK-R} & \text{STS-B} & \text{Avg.} \\ 
     \cmidrule(l){1-9} 
     
    \multicolumn{9}{l}{Encoder-only Models} \\
    \hdashline\noalign{\vskip 0.5ex}
    BERT-Base &  30.92  &59.96  &47.72  &60.35  &63.72  &58.25  &47.36 &52.65 \\
    RoBERTa-Base &53.95  &47.42  &55.87 &64.73 &63.55 &62.94 &58.40 &\textbf{58.12} \\
    \cmidrule(l){1-9}
    
    \multicolumn{9}{l}{Encoder-Decoder Models} \\
    \hdashline\noalign{\vskip 0.5ex}
    BART-Base &34.46   &52.49   &44.50   &62.51    &61.99   &57.72          &52.30   &52.28\\
    T5-Base& 37.78   &56.81   &49.37   &65.50   &64.65    &60.11    &57.52   &\textbf{55.96}
\\
    \cmidrule(l){1-9}
    
    \multicolumn{9}{l}{Decoder-only Models} \\
    \hdashline\noalign{\vskip 0.5ex}
    GPT2&     25.84	& 28.90 &	26.20 &	34.74 &	35.70 &	42.72 &	26.27 &	31.48 \\
    CLM &     27.14&	20.34&	18.73&	37.56&	27.40&	35.70&	27.97&	27.83 \\
    SimCTG &  30.32&	37.10&	31.99&	39.68&	42.73&	46.26&	25.27&	36.19 \\
    \ourmethodtok & 37.28	&37.63	&31.33	&54.78	&50.16	&48.10	&34.95	&42.03 \\ 
    \ourmethodseq & 29.66   &39.89   &34.50   &43.20  &41.99           &44.52        &25.51    &37.04\\ 
    \ourmethod & \text{37.54}	  &\text{45.23}	&\text{36.41}	 &\text{56.74}   &\text{50.30}   &\text{51.52}    &\text{39.49}   &\textbf{45.32} \\
    \bottomrule
    \end{tabular}
}
    \caption{Spearman rank correlation between the cosine similarity of sentence pairs and the human-annotated similarity scores.
    }
    \label{tab:STS_encoder_decoder}
  \end{subtable}

  \bigskip
  \bigskip
   
  \begin{subtable}{1\textwidth}
    \resizebox{\linewidth}{!} {%
    \centering
    \begin{tabular}{l c c c  c c c  c c c  c}
    \toprule
    \multirow{2}{*}{Model} & \multicolumn{3}{c}{Ruby}  & \multicolumn{3}{c}{Python} & \multicolumn{3}{c}{Java} & \multirow{2}{*}{Avg.} \\ 
    \cmidrule(lr){2-4} 
    \cmidrule(lr){5-7}
    \cmidrule(lr){8-10}
    & Ruby & Python & Java & Ruby & Python & Java & Ruby & Python & Java & \\ 
    \midrule
    \multicolumn{11}{l}{Encoder-only Models} \\
    \hdashline\noalign{\vskip 0.5ex}
    CodeBERT         & 13.55 & 3.18 & 0.71 & 3.12 & 14.39 & 0.96 & 0.55 & 0.42 & 7.62 & 4.94 \\
    GraphCodeBERT    & 17.01 & 9.29 & 6.38 & 5.01 & 19.34 & 6.92 & 1.77 & 3.50 & 13.31 & \textbf{9.17}  \\
    \midrule
    \multicolumn{11}{l}{Encoder-Decoder Models} \\
    \hdashline\noalign{\vskip 0.5ex}
    PLBART          & 18.60 & 10.76 & 1.90 & 8.27 & 19.55 & 1.98 & 1.47 & 1.27 & 10.41 & \textbf{8.25} \\
    CodeT5-base     & 18.22 & 10.02 & 1.81 & 8.74 & 17.83 & 1.58 & 1.13 & 0.81 & 10.18 & 7.81 \\
    \midrule
    \multicolumn{11}{l}{Decoder-only Models} \\
    \hdashline\noalign{\vskip 0.5ex}
    CodeGen         & 16.18 & 5.90 & 0.52 & 2.66 & 18.11 & 0.36 & 1.61 & 1.65 & 10.16 & 6.35 \\
    CLM             & 16.36 & 6.67 & 0.80 & 3.07 & 15.72 & 0.46 & 1.41 & 2.11 & 10.25 & 6.32 \\
    SimCTG          & 17.66	& 7.19 & 1.94 & 7.63 & 18.31 & 1.78 & 1.63 & 2.32 & 10.83 & 7.70 \\
    \ourmethodtok         & 18.02 & 7.84 & 2.51 & 8.76 & {20.46} & 2.48 & {1.91} & {2.58} & {11.43} & \textbf{8.44} \\
    \ourmethodseq         & 16.76	& 5.45 & 1.06 & 7.40 & 16.74 & 1.41 & 1.55 & 2.25 & 10.23 & 6.98 \\
    \ourmethod             & 17.90 & 7.78 & {2.56} & {9.05} & 19.74 & {2.64} & 1.90 & 2.50 & 11.32 & 8.38 \\
    \bottomrule
    \end{tabular}
    }
    \caption{
    MAP score (\%) of the zero-shot code-to-code search task. The language names mentioned in the top two rows indicate the languages queries and candidates are written in.
    }
    \label{tab:code_to_code_extended}
\end{subtable}
\caption{\ourmethod bridges the gap between CLM and Encoder-Only / Encoder-Decoder models.}
\label{xyz}
\end{table*}

%% file: tables/appendix/code_to_code_dropout.tex
\begin{table*}[t]
    \begin{subtable}{1\textwidth}
    \resizebox{\linewidth}{!} {%
    \centering
        \begin{tabular}{r c c c  c c c c c c c}
        \toprule
        \multirow{2}{*}{Model} & \multicolumn{3}{c}{Ruby}  & \multicolumn{3}{c}{Python} & \multicolumn{3}{c}{Java} & \multirow{2}{*}{Avg.} \\ 
        \cmidrule(lr){2-4} 
        \cmidrule(lr){5-7}
        \cmidrule(lr){8-10}
        & Ruby & Python & Java & Ruby & Python & Java & Ruby & Python & Java & \\ 
        \midrule
        CLM $_{+ \text{Dropout}}$ & 18.04 & 6.47 & 1.21 & 5.52  & 18.70  & 1.18 & 1.62  & 2.35  & 11.26  & 7.37 \\
        CLM $_{- \text{Dropout}}$ & 16.36 & 6.67  & 0.8 & 3.07 & 15.72 & 0.46 & 1.41 & 2.11 & 10.25 & 6.32 \\
        \midrule
        \ourmethod$_{+ \text{Dropout}}$     & \textbf{20.09} & \textbf{8.84} & \textbf{3.66} & \textbf{9.25} & \textbf{22.39} & \textbf{3.13} & \textbf{1.93} & \textbf{3.06} & \textbf{12.02} & \textbf{9.37} \\
         & 17.90 & 7.78 & {2.56} & {9.05} & 19.74 & {2.64} & 1.90 & 2.50 & 11.32 & 8.38 \\
        \bottomrule
        \end{tabular}
    }
    \caption{MAP score (\%) of zero-shot code-to-code search.}
    \label{tab:code_to_code_dropout}
    \end{subtable}
    
    \bigskip
    
    \begin{subtable}{1\textwidth}
    \resizebox{\linewidth}{!} {%
    \centering
        \begin{tabular}{rcccccccc}
        \toprule
        \text{Model}& \text{STS12} & \text{STS13} & \text{STS14} & \text{STS15} & \text{STS16} & \text{SICK-R} & \text{STS-B} & \text{Avg.} \\ 
         \cmidrule(l){1-9} 
        CLM $_{+ \text{Dropout}}$ &     27.14&	20.34&	18.73&	37.56&	27.40&	35.70&	27.97&	27.83 \\
        CLM $_{- \text{Dropout}}$& 25.60	&15.23	&13.95	&31.64	&28.13	&34.96	&26.15	&25.09 \\
        \cmidrule(l){1-9} 
        \ourmethod$_{+ \text{Dropout}}$ & 37.54	  &\textbf{45.23}	&\textbf{36.41}	 &\textbf{56.74}   &\textbf{50.30}   &\textbf{51.52}    &\textbf{39.49}   &\textbf{45.32} \\
        \ourmethod$_{- \text{Dropout}}$& \textbf{38.22}	&40.15	&33.57	&53.16	&45.35	&47.47	&36.10	&42.00 \\
        \bottomrule
        \end{tabular}
    }
    \caption{Spearman rank correlations between the cosine similarity of sentence representation pairs and the ground truth similarity scores for STS benchmarks.}
    \label{tab:STS_compare_dropout}
    \end{subtable}
    \caption{
    Discriminative task performances with ($_{+ \text{Dropout}}$) and without ($_{- \text{Dropout}}$) Dropout augmentation applied to CLM and \ourmethod. We apply Dropout (0.1) to all the layers of the models.
    }
    \label{tab:discriminative_tasks_dropout}
\end{table*}

%% file: tables/appendix/human_eval_dropout.tex
\begin{table*}[!htbp]
    \begin{subtable}{1\textwidth}
    \centering
        \begin{tabular}{r c c c c}
        \toprule
        \multirow{2}{*}{Model} & \multicolumn{2}{c}{Pass@k}  & \multicolumn{2}{c}{Ranked Pass@k} \\
        \cmidrule(lr){2-3} 
        \cmidrule(lr){4-5}
        & k=1 & k=5 & k=1 & k=5 \\ 
        \midrule
        CLM $_{+ \text{Dropout}}$     & 12.65 & 15.54 & 13.42 \textsubscript{(+0.77)} & 16.46 \textsubscript{(+0.92)} \\ 
        CLM $_{- \text{Dropout}}$     & 13.42 & 18.08 & 15.38 \textsubscript{(+1.96)} & 18.29 \textsubscript{(+0.21)} \\
        \midrule
        \ourmethod$_{+ \text{Dropout}}$     & 13.19 & 15.92 & 13.41 \textsubscript{(+0.22)} & 16.46 \textsubscript{(+3.05)} \\ 
        \ourmethod$_{- \text{Dropout}}$     & \textbf{14.63} & \textbf{18.83} & \textbf{17.07} \textsubscript{(+2.44)} & \textbf{18.90} \textsubscript{(+0.07)} \\
        \bottomrule
        \end{tabular}
    \caption{
    Evaluation results on the HumanEval benchmark. The numbers in the subscript indicate the difference between ranked pass@k and pass@k accuracy.
    }
    \label{tab:humaneval_dropout}
    \end{subtable}
    
    \bigskip
    \bigskip
    
    \begin{subtable}{1\textwidth}
    \resizebox{0.99\textwidth}{!}{
    \centering
        \begin{tabular}{rcccccccc}
        \toprule
        & CLM $_{- \text{Dropout}}$&  CLM $_{+ \text{Dropout}}$& \ourmethod$_{- \text{Dropout}}$ & \ourmethod$_{+ \text{Dropout}}$  \\ 
         \cmidrule(l){1-9} 
        Perplexity &\textbf{21.86}  &22.48 & 22.07 &23.01 \\
        \bottomrule
        \end{tabular}
    }
    \caption{Perplexity of continually trained GPT-2 on the test set of WikiText-103. 
    }
    \label{tab:dropout_hurts_perplexity}
    \end{subtable}
    
    \caption{
    Generation task performances with ($_{+ \text{Dropout}}$) and without ($_{- \text{Dropout}}$) Dropout augmentation applied to CLM and \ourmethod. We apply Dropout (0.1) to all the layers of the models. 
    }
    \label{tab:generative_tasks_dropout}
\end{table*}

%% file: figures/code_search_ex.tex








\begin{figure*}[t]
\centering
\begin{tabular}{l}

\hline
Query Program in Python\\
\hdashline\noalign{\vskip 0.5ex}

\begin{adjustbox}{valign=t,minipage=0.95\linewidth}
\begin{tabular}{l}
\lstset{escapechar=@,style=CustomPy}
\begin{lstlisting}
import math
h,w=map(int, input().split())
if h%3==0 or w%3==0:
    print(0)
else:
    x,y=max(h,w),min(h,w)
    ans=y
    for hi in range(1,h):
        M=max(hi*w,(h-hi)*((w+1)//2),(h-hi)*(w//2))
        m=min(hi*w,(h-hi)*((w+1)//2),(h-hi)*(w//2))
        ans=min(ans,M-m)
    for wi in range(1,w):
        M=max(wi*h,(w-wi)*((h+1)//2),(w-wi)*(h//2))
        m=min(wi*h,(w-wi)*((h+1)//2),(w-wi)*(h//2))        
        ans=min(ans,M-m)
    print(ans)
\end{lstlisting}
\end{tabular}
\end{adjustbox}

\smallskip\\
\hline
Relevant Program in Python \\
\hdashline\noalign{\vskip 0.5ex}

\begin{adjustbox}{valign=t,minipage=0.95\textwidth}
\begin{tabular}{l}
\lstset{escapechar=@,style=CustomPy}
\begin{lstlisting}
def solve(H,W):
    p1 = [H//3*W, (H-H//3)*(W//2), (H-H//3)*(W-W//2)]
    p2 = [ceil(H/3)*W, (H-ceil(H/3))*(W//2), (H-ceil(H/3))*(W-W//2)]
    S1 = max(p1)-min(p1)
    S2 = max(p2)-min(p2)
    S3 = 0 if H%3==0 else W
    return min(S1,S2,S3)

from math import ceil
H, W = map(int, input().split())
print(min(solve(H,W), solve(W,H)))
\end{lstlisting}
\end{tabular}
\end{adjustbox}

\smallskip\\
\hline
Relevant Program in Java \\
\hdashline\noalign{\vskip 0.5ex}

\begin{adjustbox}{valign=t,minipage=0.90\textwidth}
\begin{tabular}{l}
\lstset{escapechar=@,style=CustomJava}
\begin{lstlisting}
import java.util.*;
public class Main {
	public static void main(String[] args){
		Scanner sc = new Scanner(System.in);
		long w = sc.nextInt();
		long h = sc.nextInt();
		if(w%3==0 || h%3==0)
			System.out.println(0);
		else
			System.out.println(Math.min(solve(w, h), solve(h, w)));
	}
	static long solve(long w, long h){
		long min = Long.MAX_VALUE;
		for(int i=1;i<h;i++){
			long a = w*i;
			long b, c = 0, 0;
			if(w%2==0){
				b = w/2*(h-i);
				c = b;
				min = Math.min(min, Math.max(a, Math.max(b, c))-Math.min(a, Math.min(b, c)));
			}
			else if((h-i)%2==0){
				b = w*((h-i)/2);
				c = b;
				min = Math.min(min, Math.max(a, Math.max(b, c))-Math.min(a, Math.min(b, c)));
			}
			else{
				b = w*((h-i)/2);
				c = w*((h-i)/2+1);
				min = Math.min(min, Math.max(a, Math.max(b, c))-Math.min(a, Math.min(b, c)));
				b = w/2*(h-i);
				c = (w/2+1)*(h-i);
				min = Math.min(min, Math.max(a, Math.max(b, c))-Math.min(a, Math.min(b, c)));
			}
		}
		return min;
	}
}
\end{lstlisting}
\end{tabular}
\end{adjustbox}

\smallskip\\
\hline
\end{tabular}
\caption{An example of a query and relevant documents in code-to-code search task, where both query and candidates are complete programs that solve a programming problem. In this search task, each program is used as a query and retrieve all programs (from a collection of 11,744/15,594/23,530 programs in Ruby, Python, and Java, respectively) that solve the same problem.}
\label{fig:code_to_code_search}
\end{figure*}

%% file: figures/humaneval.tex
\begin{figure*}[ht]
\centering

\begin{tabular}{l}

\hline
Prompt (function signature and docstring) \\
\hdashline\noalign{\vskip 0.5ex}

\begin{adjustbox}{valign=t,minipage=0.95\textwidth}
\begin{tabular}{l}
\lstset{escapechar=@,style=CustomPy}
\begin{lstlisting}
from typing import List

def has_close_elements(numbers: List[float], threshold: float) -> bool:
    """ Check if in given list of numbers, are any two numbers closer to each other than
    given threshold.
    >>> has_close_elements([1.0, 2.0, 3.0], 0.5)
    False
    >>> has_close_elements([1.0, 2.8, 3.0, 4.0, 5.0, 2.0], 0.3)
    True
    """
\end{lstlisting}
\end{tabular}
\end{adjustbox}

\smallskip\\
\hline
Unit tests \\
\hdashline\noalign{\vskip 0.5ex}

\begin{adjustbox}{valign=t,minipage=0.95\textwidth}
\begin{tabular}{l}
\lstset{escapechar=@,style=CustomPy}
\begin{lstlisting}
def check(candidate):
    assert candidate([1.0, 2.0, 3.9, 4.0, 5.0, 2.2], 0.3) == True
    assert candidate([1.0, 2.0, 3.9, 4.0, 5.0, 2.2], 0.05) == False
    assert candidate([1.0, 2.0, 5.9, 4.0, 5.0], 0.95) == True
    assert candidate([1.0, 2.0, 5.9, 4.0, 5.0], 0.8) == False
    assert candidate([1.0, 2.0, 3.0, 4.0, 5.0, 2.0], 0.1) == True
    assert candidate([1.1, 2.2, 3.1, 4.1, 5.1], 1.0) == True
    assert candidate([1.1, 2.2, 3.1, 4.1, 5.1], 0.5) == False

check(has_close_elements)

\end{lstlisting}
\end{tabular}
\end{adjustbox}

\smallskip\\
\hline
Completion 1 (passed; mean\_logp: -0.1146) \\
\hdashline\noalign{\vskip 0.5ex}

\begin{adjustbox}{valign=t,minipage=0.95\textwidth}
\begin{tabular}{l}
\lstset{escapechar=@,style=CustomPy}
\begin{lstlisting}
    for i in range(len(numbers) - 1):
        for j in range(i + 1, len(numbers)):
            if abs(numbers[i] - numbers[j]) < threshold:
                return True
    return False
\end{lstlisting}
\end{tabular}
\end{adjustbox}

\smallskip\\
\hline
Completion 2 (failed; mean\_logp: -0.1211) \\
\hdashline\noalign{\vskip 0.5ex}

\begin{adjustbox}{valign=t,minipage=0.95\textwidth}
\begin{tabular}{l}
\lstset{escapechar=@,style=CustomPy}
\begin{lstlisting}
    for i in range(len(numbers) - 1):
        for j in range(i + 1, len(numbers)):
            if abs(numbers[i] - numbers[j]) > threshold:
                return True
    return False
\end{lstlisting}
\end{tabular}
\end{adjustbox}

\smallskip\\
\hline
Completion 3 (failed; mean\_logp: -0.1598) \\
\hdashline\noalign{\vskip 0.5ex}

\begin{adjustbox}{valign=t,minipage=0.95\textwidth}
\begin{tabular}{l}
\lstset{escapechar=@,style=CustomPy}
\begin{lstlisting}
    for i in range(len(numbers) - 1):
        if abs(numbers[i] - numbers[i + 1]) > threshold:
            return True
    return False
\end{lstlisting}
\end{tabular}
\end{adjustbox}

\smallskip\\
\hline
Completion 4 (failed; mean\_logp: -0.1607) \\
\hdashline\noalign{\vskip 0.5ex}

\begin{adjustbox}{valign=t,minipage=0.95\textwidth}
\begin{tabular}{l}
\lstset{escapechar=@,style=CustomPy}
\begin{lstlisting}
    for i in range(len(numbers) - 1):
        if abs(numbers[i] - numbers[i + 1]) > threshold:
            return False
    return True
\end{lstlisting}
\end{tabular}
\end{adjustbox}

\smallskip\\
\hline
Completion 5 (failed; mean\_logp: -0.3904) \\
\hdashline\noalign{\vskip 0.5ex}

\begin{adjustbox}{valign=t,minipage=0.95\textwidth}
\begin{tabular}{l}
\lstset{escapechar=@,style=CustomPy}
\begin{lstlisting}
    return abs(sum(numbers) - sum(numbers) / len(numbers)) < threshold
\end{lstlisting}
\end{tabular}
\end{adjustbox}

\smallskip
\\ 
\hline
\end{tabular}
\caption{
An example of prompt from HumanEval and five unique code completions generated by the \ourmethod model. We rank them based on the mean\_logp scores. Considering these 5 completions and one of the passes unit tests, pass@1 is 0.2 while ranked pass@1 is 1.0.
}
\label{fig:humaneval_ex}
\end{figure*}